\documentclass[10pt,journal,compsoc]{IEEEtran}
\ifCLASSOPTIONcompsoc
  \usepackage[nocompress]{cite}
\else
  \usepackage{cite}
\fi
\usepackage{latexsym,bm,amsmath,amssymb,amsfonts,amsthm}
\usepackage{enumerate,graphicx,subfigure}
\usepackage{algorithm,algorithmic,multirow,colortbl}
\usepackage{makecell,diagbox}
\usepackage{tablefootnote,footnote}
\graphicspath{{figures/}}
\DeclareMathOperator*{\argmax}{arg\,max}
\DeclareMathOperator*{\argmin}{arg\,min}
\newcommand\mychoose[2]{\genfrac{(}{)}{0pt}{}{#1}{#2}}
\newcommand{\be}{\begin{eqnarray}}
\newcommand{\en}{\end{eqnarray}}
\newcommand{\no}{\nonumber}
\newcommand{\ben}{\begin{eqnarray*}}
\newcommand{\enn}{\end{eqnarray*}}
\newtheorem{theorem}{Theorem}
\usepackage[colorlinks,linkcolor=red,urlcolor=blue,anchorcolor=blue,citecolor=green]{hyperref}
\usepackage{bookmark}
\begin{document}

\title{Effective Deterministic Initialization for $k$-Means-Like Methods via Local Density Peaks Searching}
\author{Fengfu~Li, %~\IEEEmembership{Student~Member,~IEEE,}
        Hong~Qiao, %~\IEEEmembership{Senior~Member,~IEEE,}
        and~Bo~Zhang %,~~\IEEEmembership{Member,~IEEE}% <-this % stops a space
\IEEEcompsocitemizethanks{
\IEEEcompsocthanksitem F. Li and B. Zhang are with LSEC and Institute of Applied Mathematics, AMSS,
Chinese Academy of Sciences, Beijing 100190, China (email: b.zhang@amt.ac.cn).\protect\\
\IEEEcompsocthanksitem H. Qiao is with State Key Lab of Management and Control for Complex Systems,
Institute of Automation, Chinese Academy of Sciences, Beijing 100190, China and
the CAS Center for Excellence in Brain Science and Intelligence Technology (CEBSIT),
Shanghai 200031, China (email: hong.qiao@ia.ac.cn).\protect\\
}% <-this % stops a space
%\thanks{This work was supported in part by NNSF of China grants 61379093 and 11131006.}
}

%%%%%%%%%%%%%%%%%%%%%%%%%%%%%%%  abstract  %%%%%%%%%%%%%%%%%%%%%%%%%%%%
\IEEEtitleabstractindextext{%
\begin{abstract}
The $k$-means algorithm is a widely used clustering method in pattern recognition and machine learning due to its simplicity to implement and low time complexity. However, it has the following main drawbacks: 1) the number of clusters, $k$, needs to be provided by the user in advance, 2) it can easily reach local minima with randomly selected initial centers, 3) it is sensitive to outliers, and 4) it can only deal with well separated hyperspherical clusters. In this paper, we propose a Local Density Peaks Searching (LDPS) initialization framework to address these issues. The LDPS framework includes two basic components: one of them is the local density that characterizes the density distribution of a data set, and the other is the local distinctiveness index (LDI) which we introduce to characterize how distinctive a data point is compared with its neighbors. Based on these two components, we search for the local density peaks which are characterized with high local densities and high LDIs to deal with the first two drawbacks of $k$-means. Moreover, we detect outliers characterized with low local densities but high LDIs, and exclude them out before clustering begins. Finally, we apply the LDPS initialization framework to $k$-medoids, which is a variant of $k$-means and chooses data samples as centers, with diverse similarity measures other than the Euclidean distance to fix the last drawback of $k$-means. Combining the LDPS initialization framework with $k$-means and $k$-medoids, we obtain two novel clustering methods called LDPS-means and LDPS-medoids, respectively. Experiments on synthetic data sets verify the effectiveness of the proposed methods, especially when the ground truth of the cluster number $k$ is large. Further, experiments on several real world data sets, Handwritten Pendigits, Coil-20, Coil-100 and Olivetti Face Database, illustrate that our methods give a superior performance than the analogous approaches on both estimating $k$ and unsupervised object categorization.
% The $k$-means clustering algorithm is popular but owns the following drawbacks: 1) $k$ needs to be provided in advance, 2) it can easily reach local minima, 3) it is sensitive to outliers, and 4) it can only deal with well separated hyperspherical clusters. In this paper, we address these issues by proposing a Local Density Peaks Searching (LDPS) initialization framework with two components: a local density that characterizes the density distribution of a data set, and a local distinctiveness index (LDI) which we introduce to characterize how distinctive a data point is compared with its neighbors. Based on them, we search for the local density peaks which are characterized with high local densities and high LDIs to deal with 1) and 2). Moreover, we detect outliers characterized with low local densities but high LDIs, and exclude them out before clustering begins. Finally, we apply the LDPS initialization framework to $k$-medoids with diverse similarity measures to fix the last drawback of $k$-means. Experiments on synthetic data sets verify the effectiveness of the proposed methods, especially when $k$ is relative large. On real world data sets, our methods give superior performance than analogous approaches on both estimating $k$ and unsupervised object categorization.
\end{abstract}

}

% make the title area
\maketitle

\IEEEdisplaynontitleabstractindextext
\IEEEpeerreviewmaketitle

%=======================================================
%                  introduction
%=======================================================
\section{Introduction}\label{sec:intro}

\IEEEPARstart{C}{lustering} methods are important techniques for exploratory data analysis with wide applications
ranging from data mining \cite{berkhin2006survey}, vector quantization \cite{coates2011importance}, dimension
reduction \cite{boutsidis2015randomized}, to manifold learning \cite{CanasPR12}. The aim of these methods is to partition
data points into clusters so that data in the same cluster are similar to each other while data in different clusters
are dissimilar. The approaches to achieve this aim include methods based on density estimation such as
DBSCAN \cite{ester1996density}, mean-shift clustering \cite{anand2014semi} and clustering by fast search and find of
density peaks \cite{rodriguez2014clustering}, methods that recursively find nested clusters \cite{murtagh2012algorithms},
and partitional methods based on minimizing objective functions such as $k$-means \cite{jain2010data},
$k$-medoids \cite{park2009simple}, the EM algorithm \cite{moon1996expectation} and ISODATA \cite{liu2012feature}.
For more information about clustering methods, see \cite{jain1999data,jain2010data,xu2005survey,berkhin2006survey}.

\begin{figure*}[!thb]
\centering
\subfigure[improper $k$]{\includegraphics[width=0.23\linewidth]{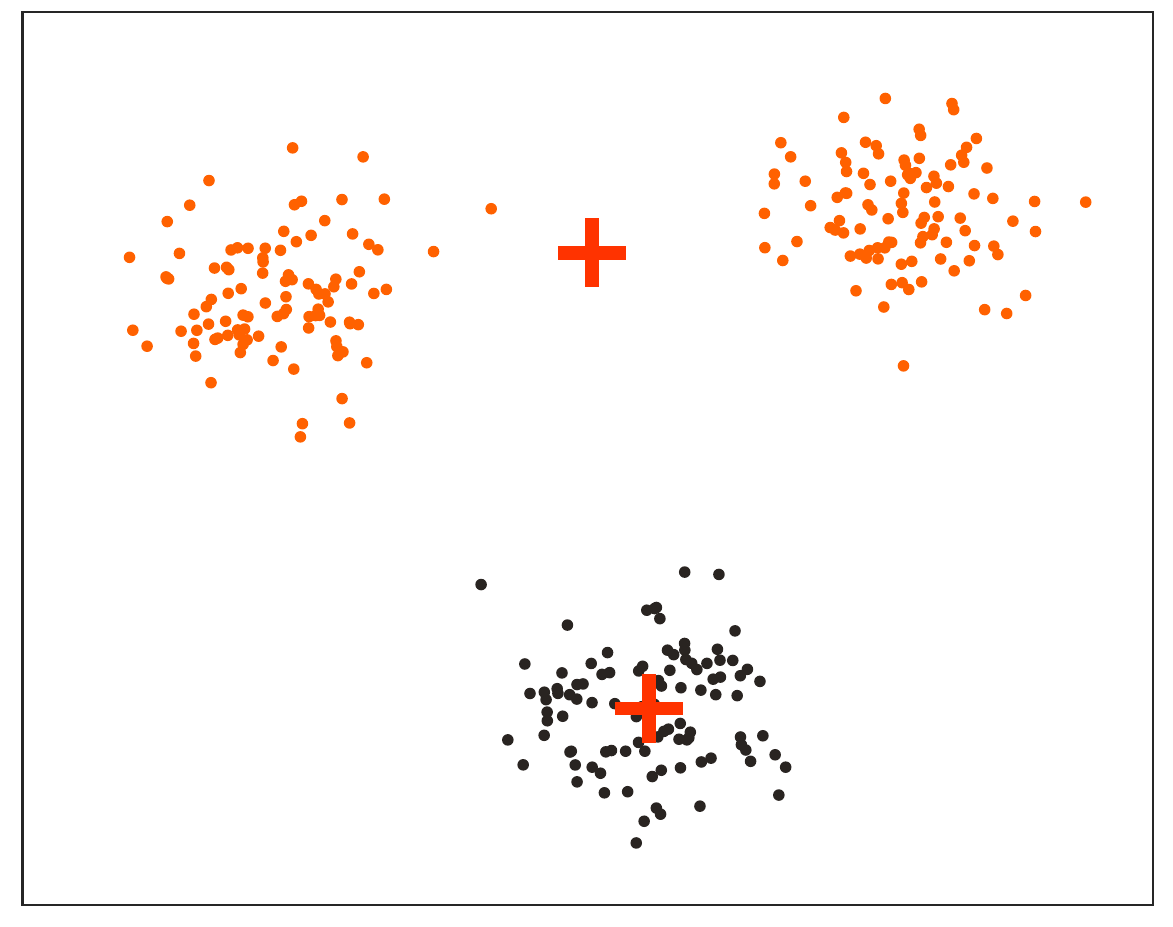}
}\quad
\subfigure[improper seeds]{\includegraphics[width=0.23\linewidth]{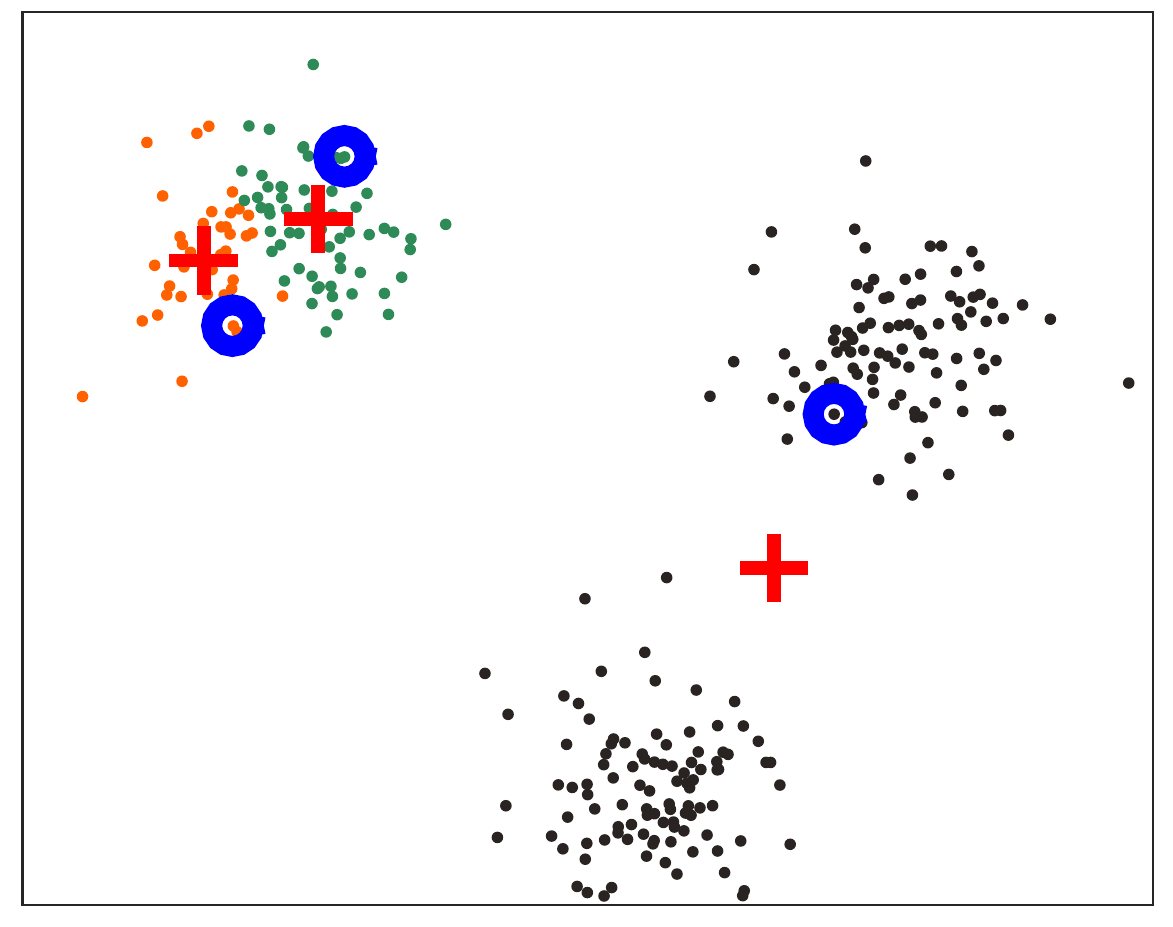}
}\quad
\subfigure[outliers effect]{\includegraphics[width=0.23\linewidth]{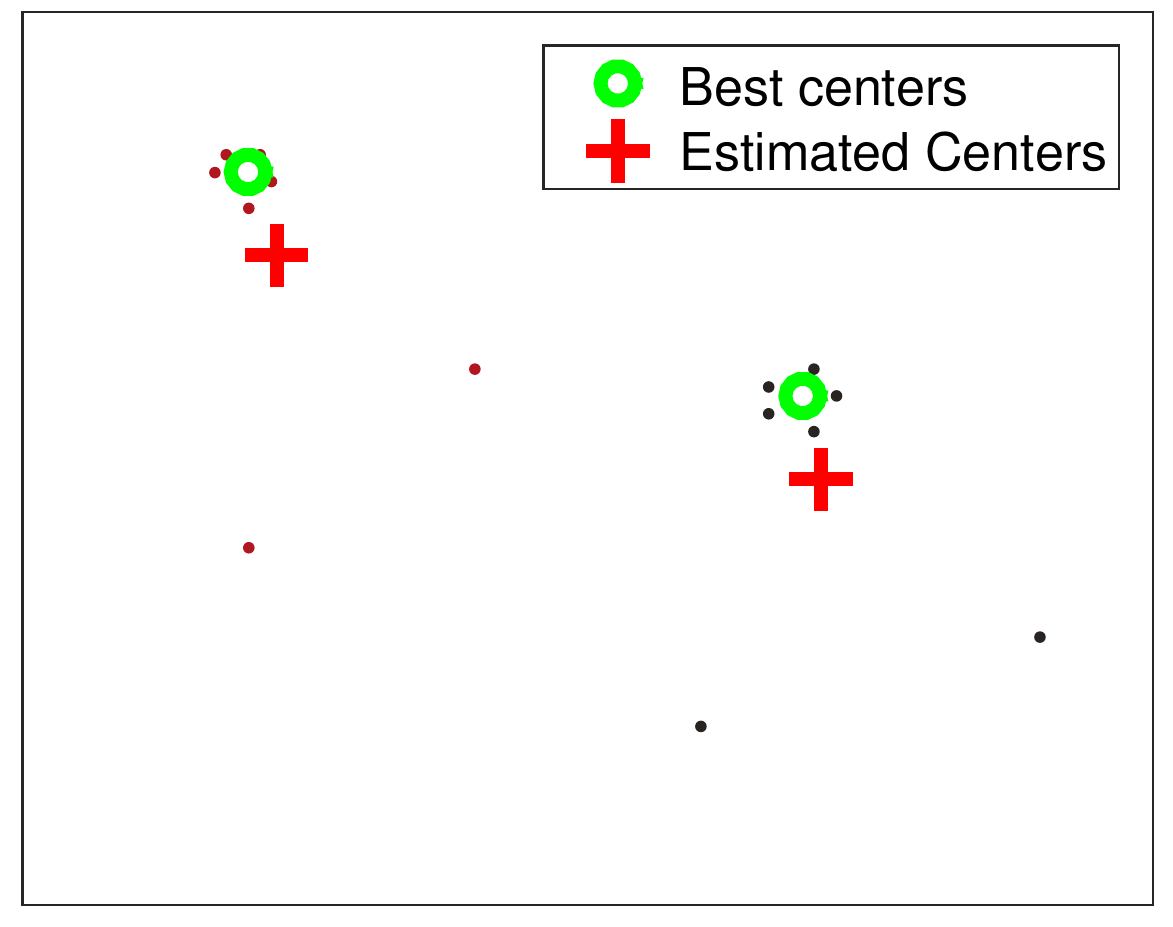}
}\quad
\subfigure[non-spherical distribution]{\includegraphics[width=0.23\linewidth]{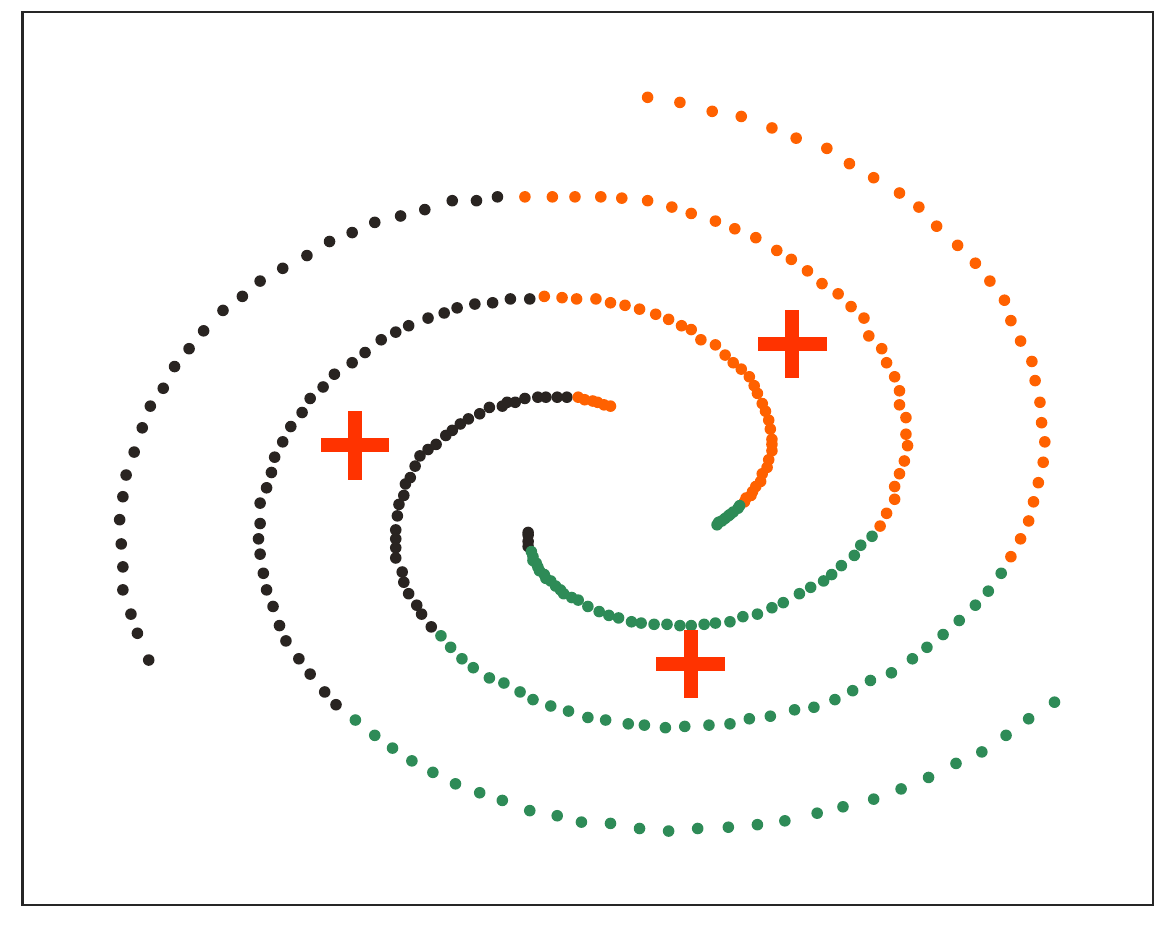}
}
\subfigure[estimated $k$]{\includegraphics[width=0.23\linewidth]{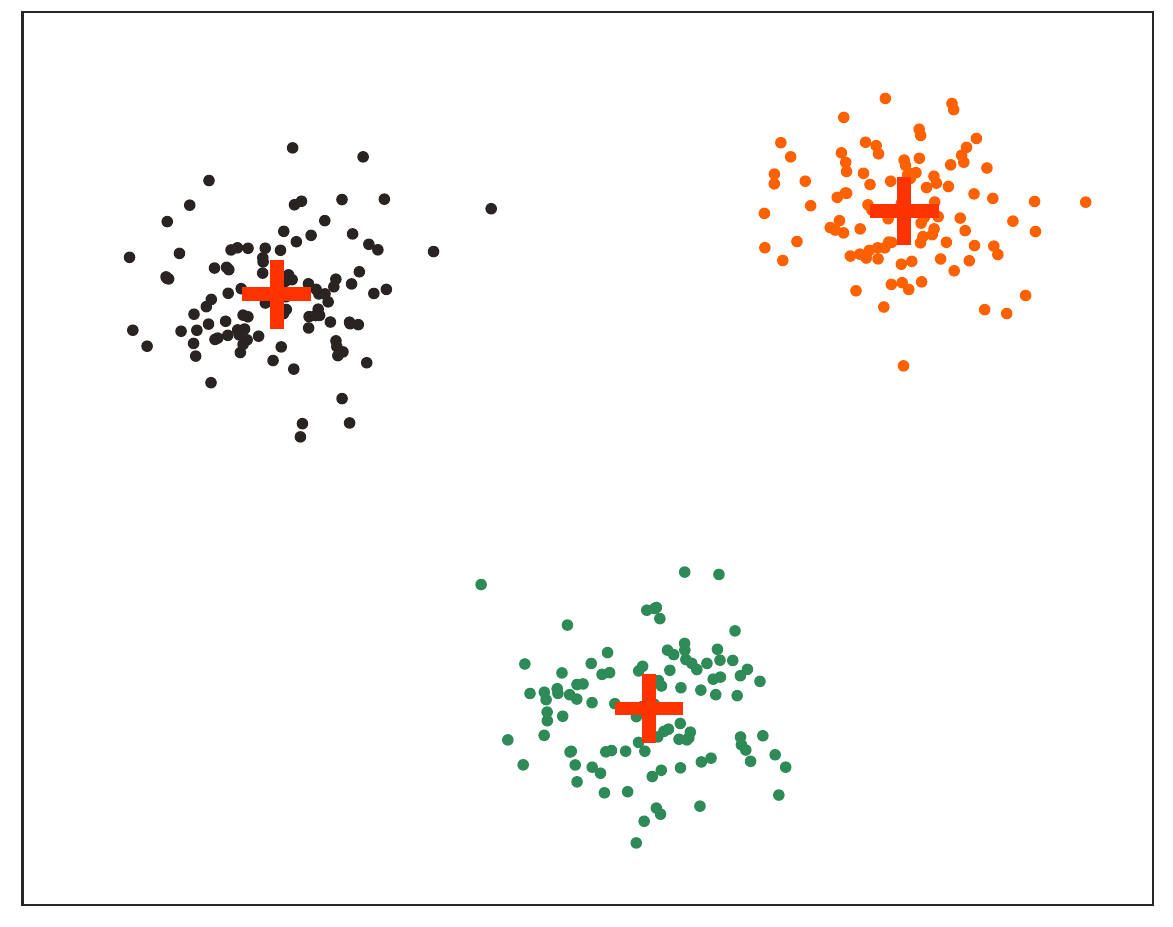}
}\quad
\subfigure[carefully selected seeds]{\includegraphics[width=0.23\linewidth]{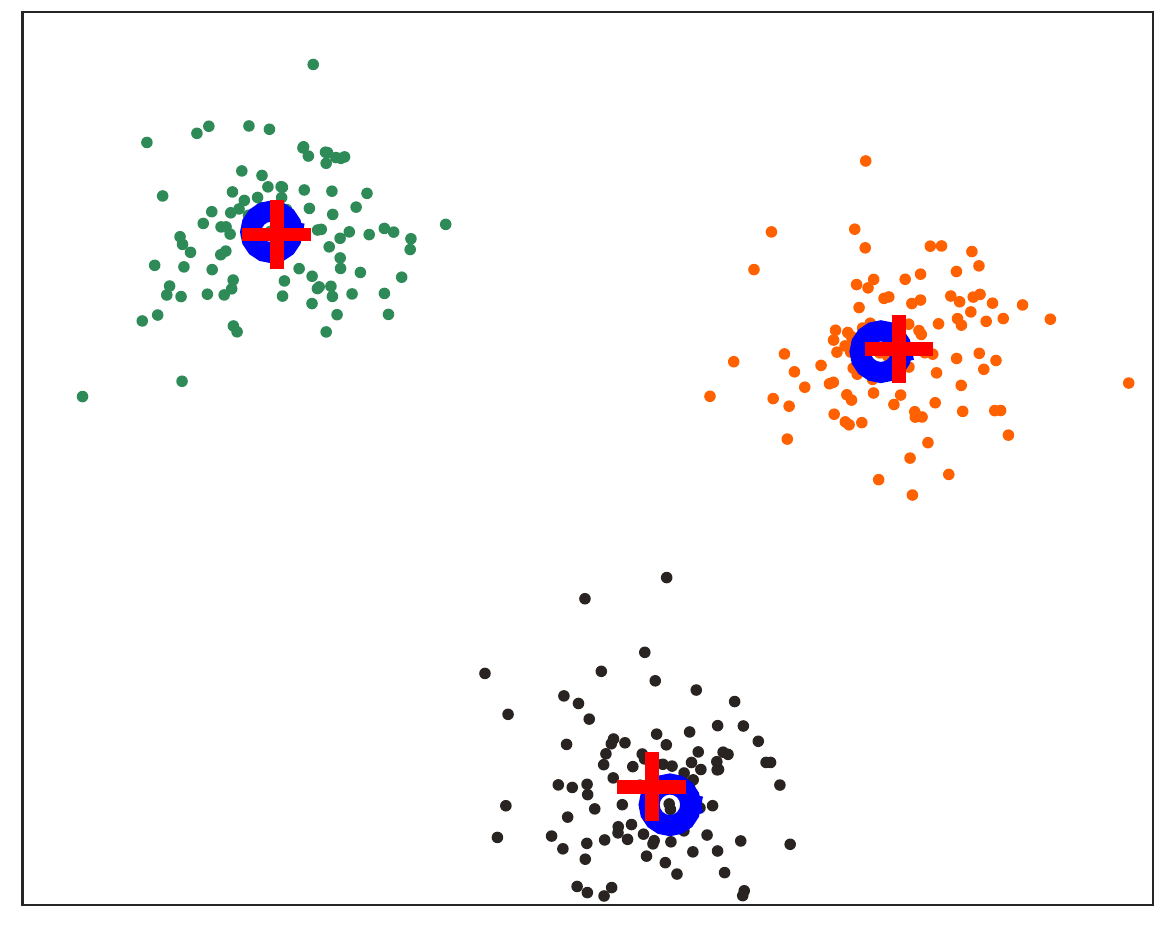}
}\quad
\subfigure[outliers detection/removal]{\includegraphics[width=0.23\linewidth]{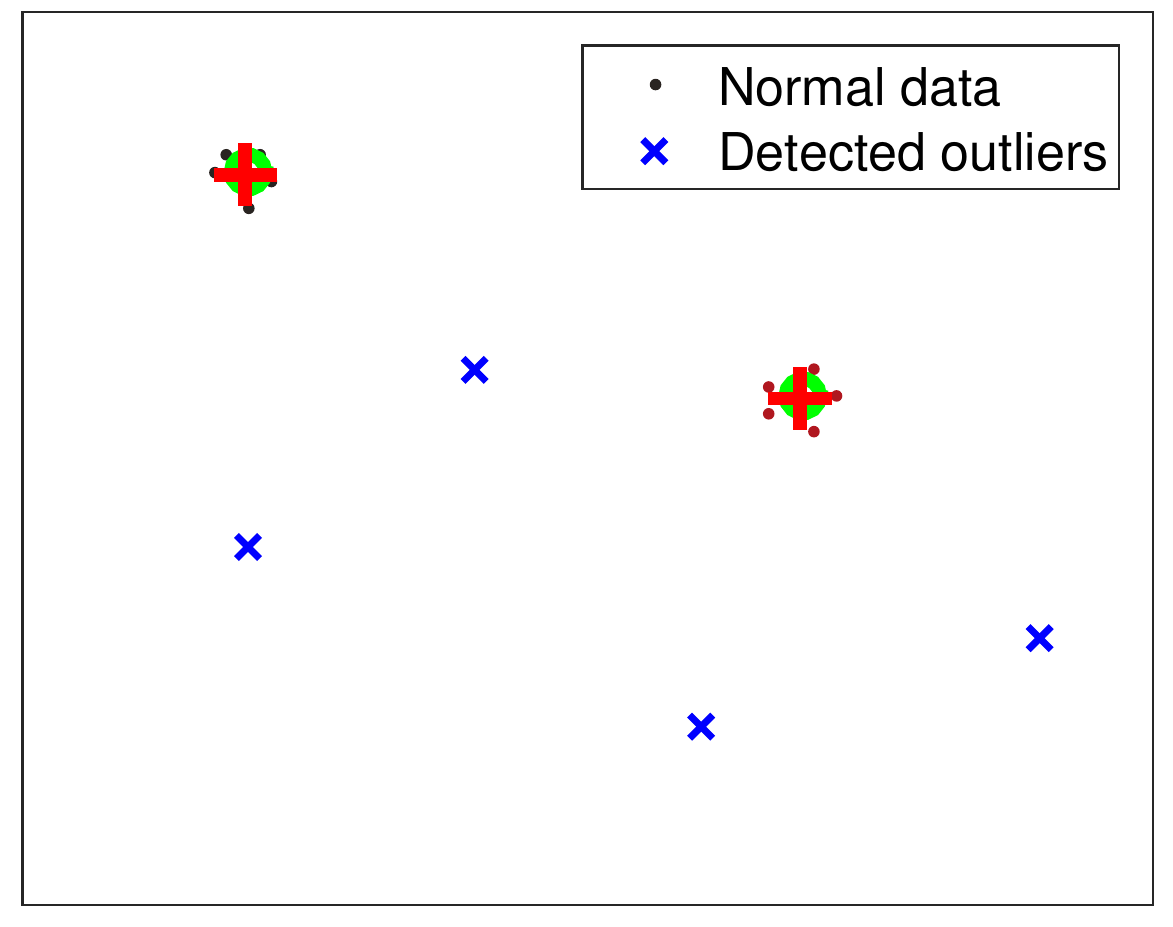}
}\quad
\subfigure[manifold-based clustering]{\includegraphics[width=0.23\linewidth]{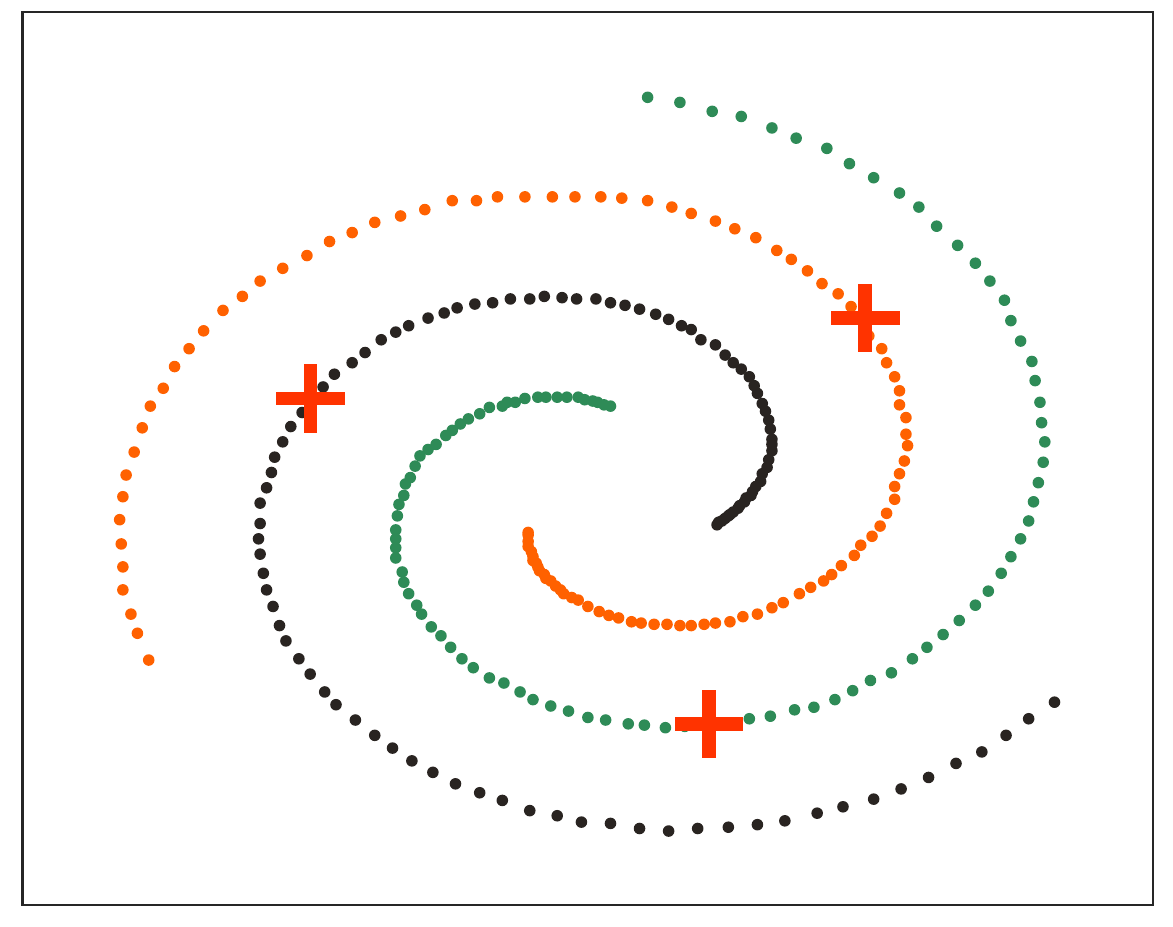}
}
\caption{Illustration of the drawbacks of $k$-means ((a)-(d)) and the solution of our methods ((e)-(h)).
First row shows the following cases for which $k$-means fails: (a) $k$ is unreasonably given,
(b) the initial seeds (marked by blue $\bm{\circ}$) are improperly selected, (c) the effect of outliers is strong, and
(d) the data distribution is non-spherical.
Second row shows our solutions: (e) an appropriate $k$ is automatically estimated, (f) the initial seeds that are
geometrically close to the center of the clusters are selected, (g) outliers are detected and removed before
clustering, and (h) the manifold-based distance is used as the dissimilarity measure instead of the squared
Euclidean distance to deal with manifold-distributed clusters.
}\label{fig:disadvantages-and-solutions}
\end{figure*}

Among partitional clustering methods, the $k$-means algorithm is probably the most popular and most widely studied
one \cite{wu2008top}. Given a set of $m$ data points $\bm{X}$ in the Euclidean space with dimensionality $p$ (thus $\bm{X}\in\mathbb{R}^{p\times m}$) and the number of clusters, $k$, the partitional clustering problem aims to determine
a set of $k$ data points in $\mathbb{R}^p$ so as to minimize the mean squared distance from each data point to its nearest
center. The $k$-means algorithm solves this problem by a simple iterative scheme for finding a local minimal solution.
It is of many advantages, including conceptually simple and easy to implement, linear time complexity and being guaranteed
to converge to a local minima (see, e.g. \cite{Selim1984,Bottou95,Celebi2013,Celebi2015}).
However, it also has the following four main drawbacks:
\begin{enumerate}[1)]
\item it requires the user to provide the cluster number $k$ in advance;
\item it is highly sensitive to the selection of initial seeds due to its gradient descent nature;
\item it is sensitive to outliers due to the use of the squared Euclidean distance;
\item it is limited to detecting compact, hyperspherical clusters that are well separated.
\end{enumerate}
Fig. \ref{fig:disadvantages-and-solutions}(a)-(d) shows the above issues on toy data sets.

\setcounter{subfigure}{0}

Many approaches have been proposed to deal with these issues individually. The first issue can be partially remedied
by extending the $k$-means algorithm with the estimation of the number of clusters. $x$-means \cite{pelleg2000x} is one
of the first such attempts that use the splitting and merging rules for the number of centers to increase and decrease
as the algorithm proceeds. $g$-means \cite{hamerly2003learning} works similarly as $x$-means except it assumes that
the clusters are generated from the Gaussian distributions. $dip$-means \cite{kalogeratos2012dip}, however, only assumes
each cluster to admit a unimodal distribution and verifies this by Hartigans' dip test \cite{hartigan1985dip}.

To address the second issue of $k$-means, a general approach is to run $k$-means repeatedly for many times and then
to choose the model with the smallest mean square error. However, this can be time-consuming when $k$ is relatively large.
To overcome the adverse effect of randomly selecting the initial seeds, many of adaptive initialization methods have been
proposed. The $k$-means{+}{+} algorithm \cite{arthur2007k} aims to avoid poor quality data partitioning during the restarts and
achieves $O(\log k)$-competitive results with the optimal clustering. The Min-Max $k$-means algorithm \cite{tzortzis2014minmax}
deals with the initialization problem of $k$-means by alternating the objective function to be weighted by the variance of
each cluster. Methods such as PCA-Part and Var-Part \cite{su2007search} use a deterministic approach based on PCA and
a variance of data to hierarchically split the data set into $k$ parts where initial seeds are selected.
For many other deterministic initialization methods, see \cite{cao09,Celebi2013,Celebi2015} and the references quoted there.

The third drawback of $k$-means, that is, its sensitivity to outliers, can be addressed by using more robust proximity measure \cite{hodge2004survey}, such as the Mahalanobis distance and the $L_1$ distance rather than the Euclidean distance.
Another remedy for this issue is to detect the outliers and then remove them before the clustering begins.
The outlier removal clustering algorithm \cite{hautamaki2005improving} uses this idea and achieves a better performance
than the original $k$-means method when dealing with overlapping clusters.
Other outliers-removing cluster algorithms can be found in \cite{Zhang03}and the references quoted there.

The last drawback of $k$-means that we are concerned with is its inability to separate non-spherical clusters.
This issue can be partially remedied by using the Mahalanobis distance to detect hyperellipsoidal clusters.
However, it is difficult to optimize the objective function of $k$-means with non-Euclidean distances.
$k$-medoids \cite{park2009simple}, as a variant of $k$-means, overcomes this difficulty by restricting the centers
to be the data samples themselves. It can be solved effectively (but slowly) by data partitioning around
medoids (PAM) \cite{kaufman2009finding}, or efficiently (but approximately optimally) by CLARA \cite{kaufman2009finding}.
In addition, the $k$-medoids algorithm makes it possible to deal with manifold-distributed data by using
neighborhood-based (dis)similarity measures. However, $k$-medoids also has the first two drawbacks of $k$-means.

In this paper, we propose a novel method named Local Density Peaks Searching (LDPS) to estimate the number of clusters
and to select high quality initial seeds for both $k$-means and $k$-medoids. A novel measure named local distinctiveness
index (LDI) is proposed to characterize how distinctive a data point is compared with its neighbors. The larger the LDI is,
the more locally distinguishable the data point is. Based on the LDI, we characterize the local density peaks by high
local densities and high LDIs. A score function is then given with the two measures to quantitatively evaluate the potential
of a data point to be a local density peak. Data points with high scores are founded and regarded as local density peaks.
By counting the number of local density peaks, a reasonable number of clusters can be obtained for further clustering
with $k$-means or $k$-medoids. In addition, the local density peaks can also be served as good initial seeds for the
$k$-means or $k$-medoids clustering algorithm. As a result, the first two drawbacks of $k$-means or $k$-medoids
are thus remedied.

In analogy with the searching of local density peaks, we characterize outliers with low local densities but high LDIs.
Another score function to quantitatively evaluate the potential of a data point to be an outlier is given.
Based on the scores, outliers can be effectively detected. To minimize the effect of outliers, we remove them
before clustering begins. Thus, the third issue of $k$-means is remedied.
Fig. \ref{fig:disadvantages-and-solutions} (e)-(h) shows the clustering results of our methods compared with
the original $k$-means algorithm.

The remainder of the paper is organized as follows. Section \ref{sec:related-works} briefly reviews some related works.
In Section \ref{sec:initialization}, we give a step by step introduction of our initialization framework for $k$-means
and $k$-medoids. Two novel clustering algorithms, called LDPS-means and LDPS-medoids, are proposed in Section \ref{sec:clustering}.
They are based on the LDPS algorithm together with the $k$-means and $k$-medoids clustering algorithms.
Section \ref{sec:performance} gives a theoretical analysis on the performance of the proposed methods.
Experiments on both synthetic and real data sets are conducted in Section \ref{sec:experiments} to evaluate the effectiveness
of the proposed methods. Final conclusions and discussions are given in Section \ref{sec:conclusion}.

%=======================================================
%                   related works
%=======================================================
\begin{figure*}[!thb]
\centering
\includegraphics[width=\linewidth]{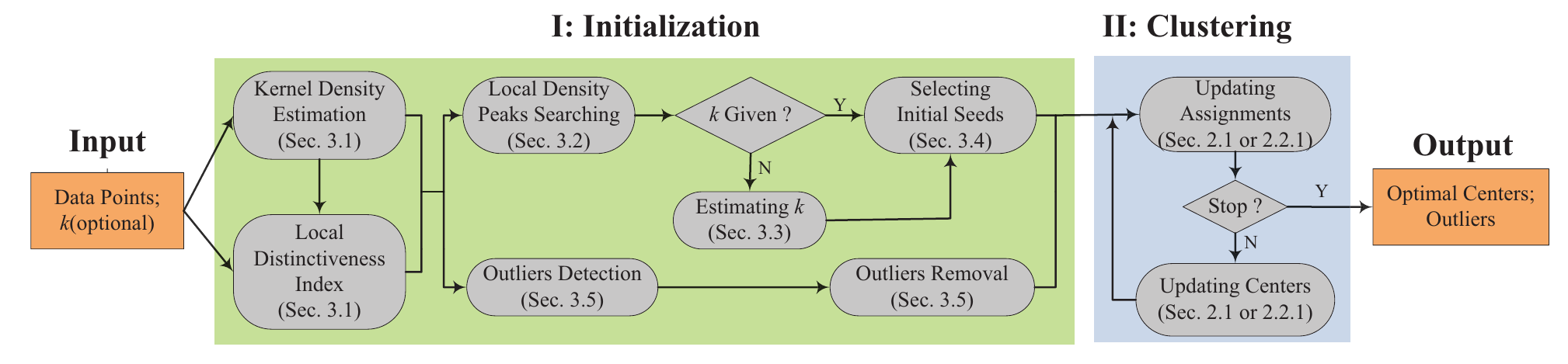}
\caption{The clustering framework as a combination of the local density and local distinctiveness index based initialization
and the $k$-means type clustering. The number $k$ of clusters is an optional input. The main part of the framework is
the initialization stage which serves as estimating $k$, selecting initial seeds and detecting/removing outliers
for the clustering stage. The clustering stage assists to find an optimized solution. Note that Y=yes, N=no.
}\label{fig:framework}
\end{figure*}

\section{Related works}\label{sec:related-works}

\subsection{The \texorpdfstring{$k$}{k}-means algorithm}

Given a data set $\bm{X}=\{\bm{x}^{(i)}\in\mathbb{R}^p|i=1,2,\ldots,m\}$, the $k$-means algorithm aims to minimize
the Sum of Squared Error (SSE) with $k$ cluster centers $\bm{C}=\{\bm{c}^{(j)}\in\mathbb{R}^p|j=1,2,\ldots,k\}$:
\begin{equation}\label{eq:sse}
\bm{C}^{*}=\mathop{\arg\min}\limits_{\bm{C}}\;\text{SSE}(\bm{X}|\bm{C})
=\sum_{i=1}^{m}\sum_{j=1}^{k}s_{ij}\,d(\bm{x}^{(i)},\bm{c}^{(j)})
\end{equation}
where $\bm{S}=\{s_{ij}|i=1,2,\ldots,m; j=1,2,\ldots,k\}$ is the assignment index set, $s_{ij}=1$ if $\bm{x}^{(i)}$
is assigned to the $j$th cluster and $s_{ij}=0$ if otherwise, $d(\bm{x},\bm{c})$ is a dissimilarity measure defined by
the Euclidean distance $d(\bm{x},\bm{c}) = \|\bm{x}-\bm{c}\|_2^2$.
To solve (\ref{eq:sse}), the $k$-means algorithm starts from a set of $k$ randomly selected initial seeds $\bm{C}^{(0)}$
and iteratively updates the assignment index set $\bm{S}$ with
\begin{equation}\label{eq:assignment}
s_{ij}=\begin{cases}
  1, & \text{if~} j = \mathop{\argmin}\limits_{1\leq l\leq k} d(\bm{x}^{(i)},\bm{c}^{(l)});  \\
  0, & \text{otherwise}
 \end{cases}
\end{equation}
and the cluster centers $\bm{C}$ with
\begin{equation}\label{eq:update-kmeans}
\bm{c}^{(j)} = \frac{1}{m_j}  \sum_{i=1}^{m} s_{ij}\bm{x}^{(i)}.
\end{equation}
Here, $m_j=\sum_i s_{ij}$ is the number of points that are assigned to the $j$th cluster. The update procedure stops
when $\bm{S}$ has no change or SSE changes very little between the current and previous iterations.
Finally, the algorithm is guaranteed to converge \cite{Selim1984} at a quadratic rate \cite{Bottou95}
to a local minima of the SSE, denoted as $\text{SSE}^*$.

\subsection{Variants of the \texorpdfstring{$k$}{k}-means algorithm}

\subsubsection{\texorpdfstring{$k$}{k}-medoids}

$k$-medoids \cite{park2009simple} has the same objective function (\ref{eq:sse}) as $k$-means. However,
it selects data samples as centers (also called \textit{medoids}), and the pairwise dissimilarity measure
$d(\bm{x},\bm{c})$ is no longer restricted to the square of the Euclidean distance. This leads to a slightly
different but more general procedure in updating the centers.
Specifically, if we denote the indices of the points in the $j$th cluster as $\bm{I}^{(j)}$$=$$\{i|s_{ij}=1\}$,
then $k$-medoids updates the center indices as
\be\label{eq:update-kmedoids}
\bm{I}^c_j=\argmin_{i\in\bm{I}^{(j)}}\sum_{l\in \bm{I}^{(j)}} d(\bm{x}^{(l)}, \bm{x}^{(i)})
\en
where $\bm{I}^c_j$ denotes the index of the $j$th center, that is, $\bm{c}^{(j)}=\bm{x}^{(\bm{I}^c_j)}$.
The assignments in $k$-medoids is updated in the same way as in $k$-means.

Compared with $k$-means, $k$-medoids is capable to deal with diverse data distributions due to its better flexibility
of choosing dissimilarity measures. For example, $k$-means fails to discover the underlying manifold structure of the
manifold-distributed data, while $k$-medoids may be able to do so with manifold-based dissimilarity measures.

\subsubsection{\texorpdfstring{$x$}{x}-means and \texorpdfstring{$dip$}{dip}-means}

$x$-means \cite{pelleg2000x} is an extension of $k$-means with the estimation of the number $k$ of clusters.
It uses the splitting and merging rules for the number of centers to increase and decrease as the algorithm proceeds.
During the process, the Bayesian Information Criterion (BIC) \cite{kass1995reference} is applied to score the modal.
The BIC score is defined as follows:
\ben
\text{BIC}(M|\bm{X})=L(\bm{X}|M)-\frac{l}{2}\log(m),
\enn
where $L(\bm{X}|M)$ is the log-likelihood of the data set $\bm{X}$ according to the modal $M$ and
$l=k(p+1)$ is the number of parameters in the modal $M$ with the dimensionality $p$ and $k$ cluster centers.
$x$-means chooses the modal with the best BIC score on the data. The BIC criterion works well only for the case
where there are a plenty of data and well-separated spherical clusters.

$dip$-means \cite{kalogeratos2012dip} is another extension of $k$-means with the estimation of the number of clusters.
It assumes that each cluster admits a unimodal distribution which is verified by Hartigans' dip test.
The dip test is applied in each split candidate cluster with a score function
\ben
score_j =\begin{cases}
\frac{1}{|\bm{v}_j|}\sum_{\bm{x}\in \bm{v}_j} dip(F^{(\bm{x})}_{\alpha}),& \frac{|\bm{v}_j|}{m_j}\geq v_{thd}\\
0\text{~~~~~~~~~~~~~~~~~~~~~~~~~~~~~~~},& \text{otherwise}.
\end{cases}
\enn
Here, $\bm{v}_j$ is the set of split viewers, $\alpha$ is a statistic significant level for the dip test.
The candidate with the maximum score is split in each iteration. It works well when the data set has various
structural types. However, it would underestimate $k$ when the clusters are closely adjacent.

\subsubsection{\texorpdfstring{$k$}{k}-means++}

$k$-means++ is a popular variant of $k$-means with adaptive initialization. It randomly selects the first center
and then sequentially chooses $\bm{x}\in \bm{X}$ to be the $j$th ($2\leq j\leq k$) center with probability
${D(\bm{x})^2}/({\sum_{\bm{x}'}D(\bm{x}')^2})$, where $D(\bm{x})$ is the minimum distance from $\bm{x}$
to the closest center that we have already chosen. The capability of the method is reported to
be $O(\log k)$-competitive with the optimal clustering.

\subsection{Clustering by fast search and find of density peaks}

Clustering by fast search and find of density peaks (CFSFDP) \cite{rodriguez2014clustering} is a novel clustering method.
It characterizes a cluster center with a higher density than its neighbors and with a relatively large distance
from other points with high densities. The density $\rho_i$ of $\bm{x}^{(i)}$ is defined as
\begin{equation}\label{eq:dens-cutoff}
  \rho_i = \sum_{j=1}^{m} \chi(d_{ij}-d_c)
\end{equation}
where $\chi(z)=1$ if $z<0$ and $\chi(z)=0$ if otherwise, $d_{ij}$ is the distance between $\bm{x}^{(i)}$
and $\bm{x}^{(j)}$ and $d_c$ is a cutoff distance. Intuitively, $\rho_i$ equals to the number of the points
whose distance from $\bm{x}^{(i)}$ is less than $d_c$. Another measure $\delta_i^g$, which we call the
\textit{global distinctiveness index (GDI)}, is the minimum distance between $\bm{x}^{(i)}$ and
any other points with high densities:
\begin{equation}\label{eq:gdi}
  \delta^g_i = \min_{j:\rho_j > \rho_i} d_{ij}.
\end{equation}
For the point with the peak density, its GDI is defined as $\delta_i^g=\max_j d_{ij}$. Only those points with
relatively high local densities and high GDIs are considered as cluster centers. By combining the local density
and GDI, a possible way of choosing the cluster centers is to define
\ben
 \gamma_i = \rho_i \cdot \delta_i^g
\enn
and then to choose the points with high values of $\gamma$ to be the centers. After the cluster centers are determined,
each of the remaining points is assigned to the cluster if it is the nearest neighbor of the cluster center with
a higher density.

Though the CFSFDP method is simple, it is powerful to distinguish clusters with distinct shapes. In addition,
it is insensitive to outliers due to a cluster halo detection procedure. However, CFSFDP also has some disadvantages,
such as: 1) it does not give a quantitative measure of how to choose the cluster centers automatically,
2) its assignment step is not clear compared with the $k$-means and $k$-medoids algorithms,
3) it is sensitive to the parameter $d_c$, and 4) it cannot serve as an initialization method for $k$-means or
$k$-medoids when the number $k$ of the clusters is given in advance.

%=======================================================
%                   initialization
%=======================================================

\begin{figure*}[!thb]
\centering
\subfigure[The R15 data set.]{
\includegraphics[width=0.31\textwidth]{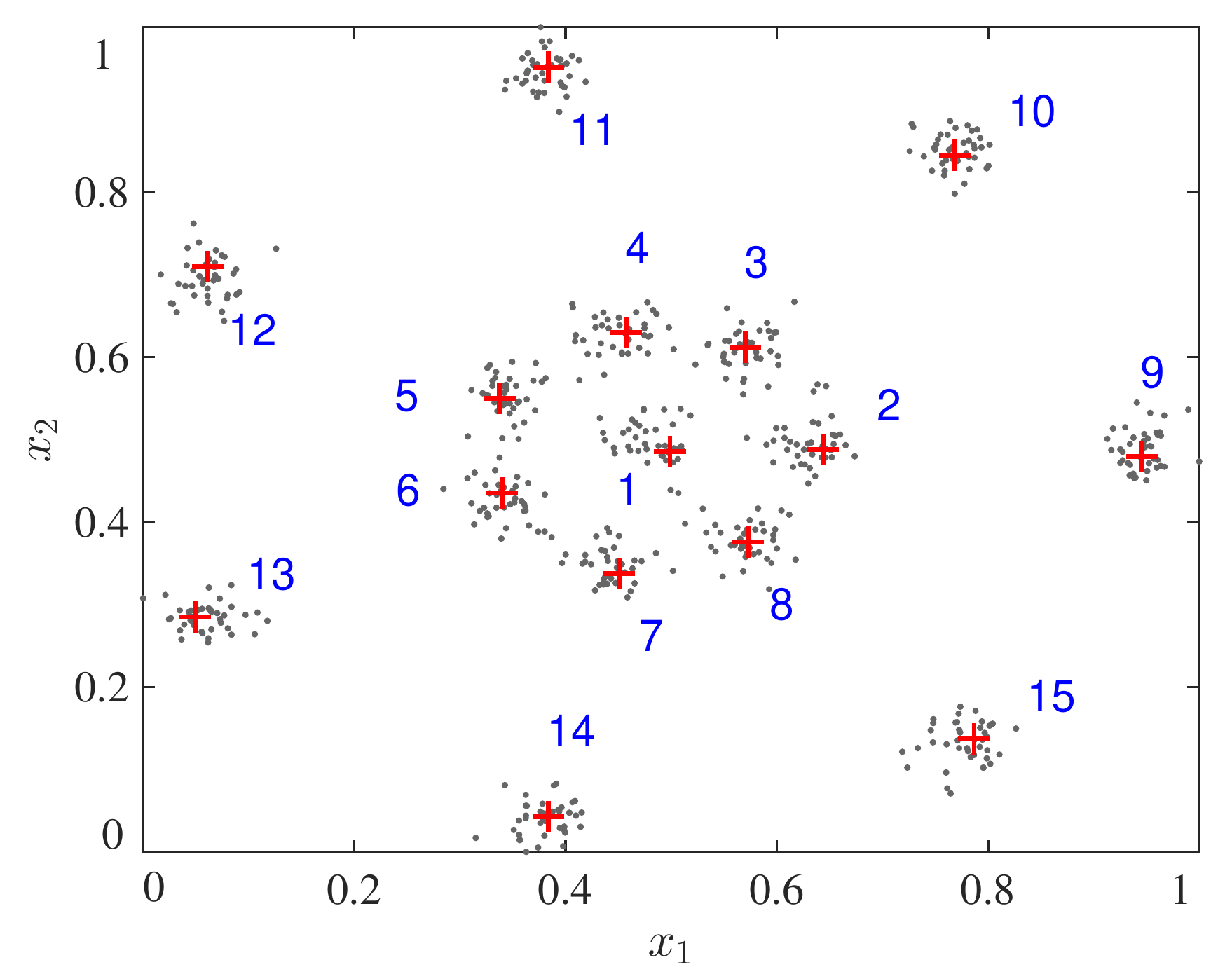}
}
\subfigure[$\rho$-$\delta^g$ graph of R15.]{
\includegraphics[width=0.31\textwidth]{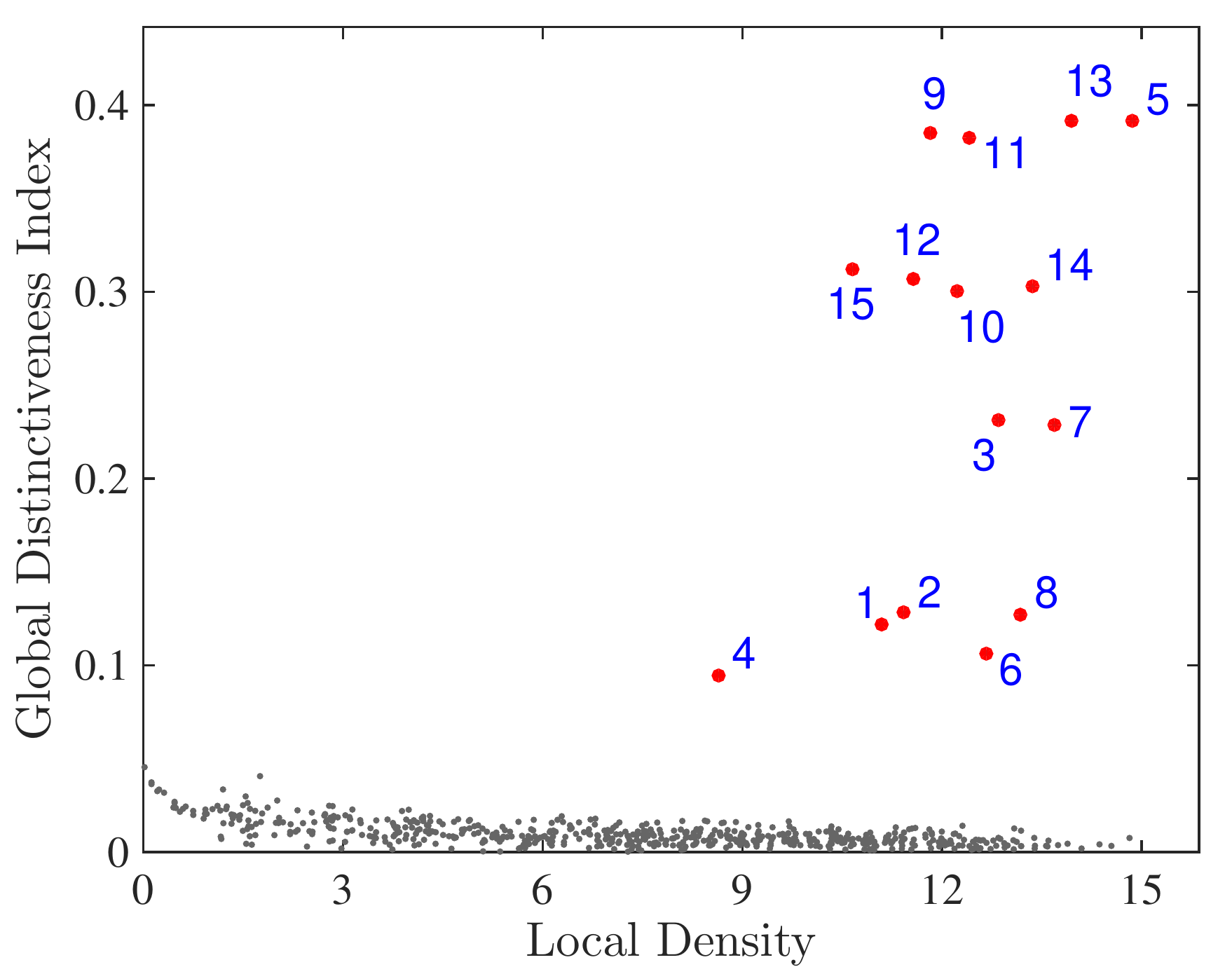}
\label{fig:R15-gdi}
}
\subfigure[$\overline{\rho}$-$\delta^l$ graph of R15.]{
\includegraphics[width=0.31\textwidth]{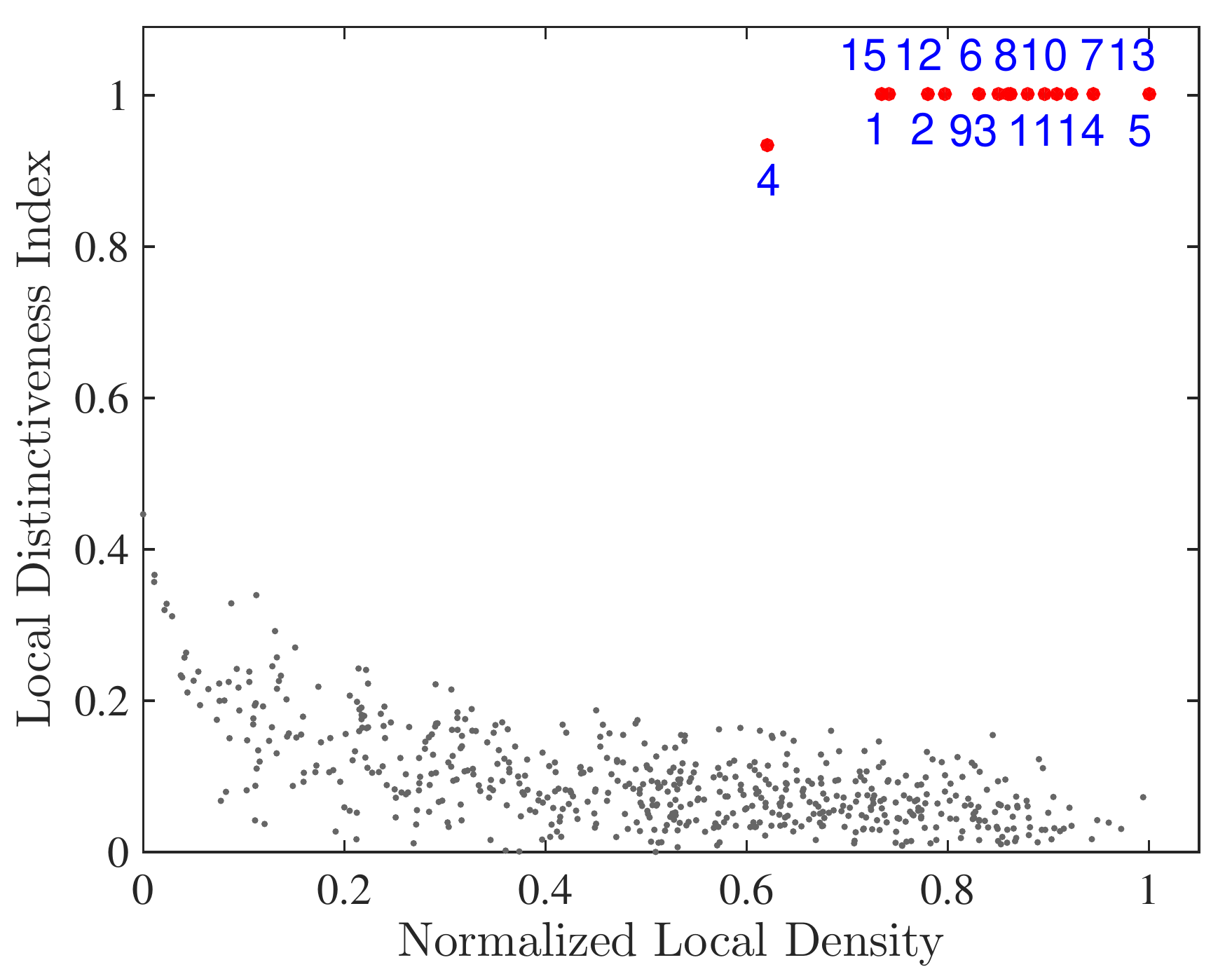}
\label{fig:R15-ldi}
}
\caption{Comparison of GDI and LDI on the R15 data set. (a) The R15 data set which owns 15 clusters.
Points marked with (red) "\textbf{+}" own highest local densities within their neighborhoods.
(b) The dominant effect of GDI. Although the 15 labeled points have similar highest local densities
in their neighborhoods, their GDIs vary a lot from 0.1 to 0.4. The GDIs of points 1, 2, 4, 6, and 8
are dominated by points 3, 5 and 7 which own relatively larger densities than the aforementioned points.
(c) The LDI wiped out the dominant effect by choosing proper $r$; where $r=0.1d^*$.
Now, labeled points almost all have largest LDIs. As a result, they are quantitatively more distinctive
than the unlabeled points. }
\label{fig:gdi-and-ldi}
\end{figure*}

\section{Local density and LDI based initialization for $k$-means-like methods}
\label{sec:initialization}

In this section, we propose an initialization framework for the $k$-means and $k$-medoids algorithms.
Fig. \ref{fig:framework} illustrates the overall diagram for the clustering framework, where Part I is
the initialization framework. We first introduce two basic measures: one is the local density and the other is
the local distinctiveness index (LDI). Based on these two measures, we propose a local density peaks searching
algorithm to find the local density peaks, which can be used to estimate the number $k$ of clusters as well as
to serve as the initial seeds for $k$-means or $k$-medoids. In addition, outliers can be detected and removed
with the help of the measures. Below is a detailed description of the proposed method.

\subsection{Local density and local distinctiveness index} \label{subsec:dens-and-ldi}

Local density characterizes the density distribution of a data set. We use the kernel density estimation
(KDE) \cite{sheather2004density} to compute the local densities.
Suppose the samples $\bm{X}$ are generated from a random distribution with a unknown density $f$.
Then KDE of $f$ at the point $\bm{x}^{(i)}$ is given by
\begin{equation}\label{eq:dens-gaussian}
\rho_{_{i}}=\hat{f}(\bm{x}^{(i)}) = \frac{1}{mh} \sum_{j=1}^{m} K(\frac{ d_{ij} }{h}),
\end{equation}
where $h$ is a smoothing parameter called \textit{bandwidth}, $d_{ij}$ is the distance between $\bm{x}^{(i)}$
and $\bm{x}^{(j)}$, and $K(z)\,(z\in\mathbb{R})$ is a kernel function satisfying that 1) $K(z)\geq 0$,
2) $\int K(z)dz=1$ and 3) $K(-z)=K(z)$. A popular choice for the kernel function is the standard Gaussian kernel:
\begin{equation}\label{eq:gaussian}
K(z) = \frac{1}{\sqrt{2\pi}}\text{exp}(-\frac{z^2}{2}).
\end{equation}

It is known that the Gaussian kernel is smooth. In addition, compared with the uniform kernel
$K(z)=[1/({b-a})]I_{\{a\leq z\leq b\}}$ which is used in \cite{rodriguez2014clustering} (see (\ref{eq:dens-cutoff})),
the Gaussian kernel has a relatively higher value when $|z|$ is small and thus keeps more local information near zero.
In what follows, we will use the Gaussian kernel.

Based on the local density, we propose a new measure called the \textit{local distinctiveness index (LDI)} to evaluate
the distinctiveness of the point $\bm{x}^{(i)}$ compared with its $r$-neighbors
$N_r(\bm{x}^{(i)})=\{\bm{x}^{(j)}\,|\,0<d_{ij}\leq r\}$. We first define the \textit{local dominating index set (LDIS)}:
\ben
\text{LDIS}(\bm{x}^{(i)})=\{j|\bm{x}^{(j)}\in N_r(\bm{x}^{(i)})\;\text{and}\;\rho_j>\rho_i\}.
\enn
Intuitively, $\text{LDIS}(\bm{x}^{(i)})$ indicates which of the points in the $r$-neighbors of $\bm{x}^{(i)}$ dominates
$\bm{x}^{(i)}$ with the local density measure. Based on LDIS, we can define LDI as follows:
\be\label{eq:ldi}
\delta_i^l=\begin{cases}
1 & \text{if}\,\, \text{LDIS}(\bm{x}^{(i)}) = \varnothing, \\
\mathop{\min}\limits_{j\in\textit{LDIS}(\bm{x}^{(i)})} d_{ij}/ r & \text{elsewise}.
\end{cases}
\en
where $\varnothing$ denotes the empty set.

With the definition (\ref{eq:ldi}), $\delta^l$ lies in $(0, 1].$ The point $\bm{x}^{(i)}$ has the biggest LDI
if its LDIS is empty, which means that $\bm{x}^{(i)}$ is not dominated by any other point, that is,
either $N_r(\bm{x}^{(i)})$ is empty or $\rho_j\leq \rho_i$ for any $j\in N_r(\bm{x}^{(i)})$.
For any other point, its LDI is computed as the minimal distance between the point and the dominating points,
divided by the local parameter $r$. When $r$ is set to be larger than $d^*=\max_{i,j}d_{ij}$,
the LDI will degenerate to the GDI since $N_r(\bm{x}^{(i)})=\bm{X}$ for any $\bm{x}^{(i)}$.
Thus, LDI is a generalization of GDI.
However, LDI characterizes the local property of the data distribution, but GDI does not give us any local information
of the data distribution. Fig. \ref{fig:gdi-and-ldi} shows the difference between GDI and LDI.
The GDI of the point with the highest density is defined as the global maximum distance between the point
and all other points, and GDIs of the other points are defined by their maximum distance to the points
with higher densities. Thus, even though two points have similar highest local densities within their neighborhoods,
their GDIs may have a big difference. We call this phenomenon as the \textit{dominant effect} of GDI
(see Fig. \ref{fig:gdi-and-ldi}(a)-(b)). Fortunately, the dominant effect of GDI can be eliminated by
using LDI with the appropriate choice of the parameter $r$ since LDI of a point is only affected by the points
within its $r$-neighborhood (see Fig. \ref{fig:gdi-and-ldi}(c)). Thus, LDI will be quantitatively more distinctive
than GDI when the number of clusters is large.

% \footnotetext{For more information about R15, see sec. \ref{sec:experiments}.}

\subsection{Local density peaks searching}  \label{subsec:ldps}

\begin{figure}[!htb]
\centering
\includegraphics[width=0.5\linewidth]{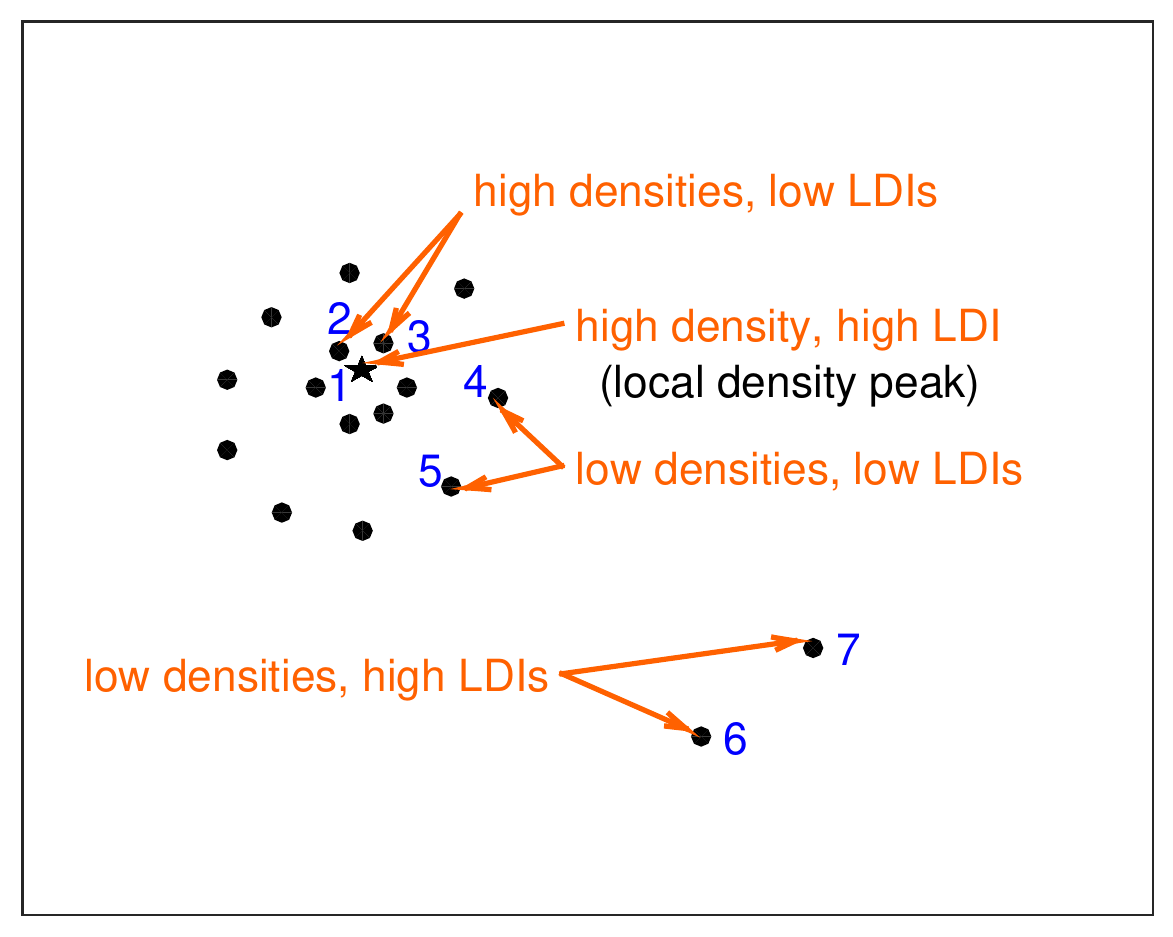}\\
\caption{Local density and LDI distribution. Point 1 is the unique density peak which has the highest local density
and highest LDI among all of the points. Points 2 and 3 have relatively large densities, but their LDIs are very small
since their LDISs include Point 1, and their distances to Point 1 are very small. Points 4-7 all have relatively low densities,
but their LDIs are different. Points 4 and 5 are relatively close to the center and thus their LDIs are small.
Points 6 and 7, however, are far away from the cluster, and as a result, their LDIs are relatively high.
}\label{fig:representatives}
\end{figure}

In the CFSFDP algorithm, a density peak is characterized by both a higher density than its neighbors and a relatively
large distance from the points with higher densities. We use the similar idea to search for the \textit{local} density peaks
by assuming that a local density peak should have a high local density and a large local distinctiveness index.
Fig. \ref{fig:representatives} gives an intuitive explanation of how this works.

To find the local density peaks, we introduce a quantitative measure $\gamma^c$ to evaluate the potential of a point
to be a local density peak. $\gamma^c$ is defined as
\begin{eqnarray}\no
\gamma^c_i & = & \left(1-\frac{1}{2}(1-\overline{\rho}_i)^2-\frac{1}{2}(1-\delta_i^l)^2\right)^2 \\ \label{eq:vc}
 &=&\left(1-\underbrace{ (1-\overline{\rho}_i)(1-\delta_i^l)}_{\text{I}}
    -\frac{1}{2}\underbrace{(\overline{\rho}_i-\delta_i^l)}_{\text{II}}{}^2\right)^2 . %\label{eq:vc-2}
\end{eqnarray}
Here, $\overline{\rho}_i=\rho_i/{\mathop{\max}\limits_i\rho_i}$ is the normalized local density.

By definition (\ref{eq:vc}), $\gamma^c$ lies in $[0,1]$ and is an increasing function of $\rho$ and $\delta^l$.
Further, the term I in the second equation in the definition (\ref{eq:vc}) is used for selecting the high local density
and high LDI points. For instance, the point with a high density ($\overline{\rho}>0.9$) and a high LDI ($\delta^l>0.9$)
will have the $\gamma^c$ value being close to $1.$
On the contrary, the point with a low local density ($\overline{\rho}<0.1$) and a low local LDI ($\delta^l<0.1$)
will have the $\gamma^c$ value being close to $0.$ The term II in the second equation in the definition (\ref{eq:vc})
is to balance the influence of the local density and LDI.
As a consequence, the points with balanced local densities and LDIs are preferred. For example,
$\gamma^c|_{\overline{\rho}=0.5,\delta^l=0.6}$ equals to $0.64,$ which is much greater than
$\gamma^c|_{\overline{\rho}=1,\delta^l=0.1}$ that equals to $0.36.$
%In addition, the square operation is applied to expand the former influences.

We now analyze the local density and LDI distribution in Fig. \ref{fig:representatives}. First,
the point which lies in the centroid of a cluster will have the highest local density and LDI
and thus is regarded as a local density peak. Secondly, points which are close to the centroid will have
high local densities and low LDIs since their LDISs all include the local density peak, and their distances
to the local density peak are small. Finally, points which are far away from the centroid will have relatively
low local densities. Quantitatively, the local density peaks in a local area will have high $\gamma^c$ values,
but the other points will have relatively much smaller $\gamma^c$ values by the definition (\ref{eq:vc}).
Thus, there should have a big gap in terms of the $\gamma^c$ value between the local density peaks and
the other points around the local density peakes. By observing the gap, the local density peaks can be found.
Based on the above discussion we propose the Local Density Peaks Searching (LDPS) algorithm which is stated
in Algorithm \ref{alg:ldps}.

\begin{algorithm}[!htb]
\caption{The LDPS Algorithm}\label{alg:ldps}
\begin{algorithmic}[1]
\REQUIRE dissimilarity matrix $\bm{D}$, bandwidth $h$ and local parameter $r$
\ENSURE $\bm{I}_{ldp}$ and $\tau^*$
\STATE Compute local densities $\bm{\rho}$ with (\ref{eq:dens-gaussian})
\STATE Compute local distinctiveness index $\bm{\delta^l}$ by (\ref{eq:ldi})
\STATE Compute $\bm{\gamma}^c$ with $\bm{\rho}$ and $\bm{\delta}^l$ by (\ref{eq:vc})
\STATE Sort $\bm{\gamma}^c$ with the descending order:
$$
  [\bm{\gamma}^{cs},\, \bm{I}^{cs}] = \text{sort}(\bm{\gamma}^c\,|\,\text{"descend order"})\,
$$
\vspace{-0.4cm}
\STATE Compute the gaps $\bm{\tau}$ (negative difference of $\bm{\gamma}^{cs}$):
$$
\tau_i=-(\Delta \bm{\gamma}^{cs})_i = \bm{\gamma}^{cs}_i - \bm{\gamma}^{cs}_{i+1}\,
$$
\vspace{-0.4cm}
\STATE Observe the biggest gap and decide the number of the local density peaks:
$$
k = \mathop{\argmax}\limits_i \tau_i\,
$$
\vspace{-0.4cm}
\STATE Search for the local density peaks with indices:
$$\bm{I}^{ldp} = \mathop{\cup}\limits_{i=1}^{k}\{\bm{I}^{cs}_i\}$$
\vspace{-0.4cm}
\STATE Compute the maximum gap of the $\gamma^c$ value between the local density peaks and the other points:
$$
\tau^*=\tau_{k}
$$
\vspace{-0.4cm}
\RETURN $I_{ldp}$ and $\tau^*$
\end{algorithmic}
\end{algorithm}

In Algorithm \ref{alg:ldps}, $\bm{\gamma}^{cs}$ is the sorted vector of $\bm{\gamma}^c$ with the descending order,
$\bm{I}^{cs}$ is the sorted index, that is, $\bm{\gamma}^{cs}_i=\bm{\gamma}^c_{\bm{I}^{cs}_i}$, and $\Delta$ is
the numerical difference operator:
$$\Delta f(z_j)=\frac{f(z_{j+1})-f(z_{j})}{z_{j+1}-z_j}.$$

\subsection{Estimating \texorpdfstring{$k$}{k} via local density peaks searching}\label{subsec:estimate-k}

$k$-means and $k$-medoids require the number $k$ of the clusters as an input. However, it is not always easy to
determine the best value of $k$ \cite{pelleg2000x}. Thus, learning $k$ is a fundamental issue for the clustering algorithms.
Here, we use the LDPS algorithm for estimating $k$ which is equal to the number of the local density peaks.

$\tau^*$ in Algorithm \ref{alg:ldps} is the minimum gap of the $\gamma^c$ value between the selected local
density peaks and the other points and thus treated as a measure of how distinctive the local density peaks are.
The bigger the value of $\tau^*$ is, the better the estimated $k$ will be. If the resulting $\tau^*$ is too small,
the procedure for estimating $k$ will fail. In this case, we set the estimated $k$ as $-1$.

Compared with $x$-means and $dip$-means which are incremental methods that use the splitting/merging rules
to estimate $k$, our method does not have to split the data set into subsets and is thus fast and more stable.
Further, CFSFDP uses the two-dimensional $\bm{\rho}$-$\bm{\delta}^g$ decision graph (see Fig. \ref{fig:R15-gdi})
together with manual help to select the cluster centers, while our method estimates $k$ quantitatively
and automatically without any manual help.

\subsection{Selecting initial seeds with local density peaks} \label{subsec:init-seeds}

Choosing appropriate initial seeds for the cluster centers is a very important initialization step
and plays an essential role for $k$-means and $k$-medoids to work properly.
Here, we assume that we have already known the true number $k$ of the clusters (either given by the user or estimated
by using the LDPS algorithm). Let us denote by $k^*$ the true number of the clusters used for the clustering algorithms.

If the true number $k^*$ of the clusters is not given in advance by the user, we use the estimated $k$ to be $k^*$.
In addition, we take the local density peaks $\{\bm{x}^{(i)}|i\in\bm{I}^{ldp}\}$ obtained by the LDPS algorithm
to be the initial seeds.
%Otherwise, the estimated $k$ may not be equal to the $k^*$ provided by the user.
In fact, we select the first $k^*$ elements with the leading $\bm{\gamma}^c$ values as the initial seeds, that is,
\begin{equation}\label{eq:initial-seeds}
\bm{I}^{s}=\mathop{\cup}\limits_{i=1}^{~~k^*}\{\bm{I}^{cs}_i\},
\end{equation}
where $\bm{I}^{s}$ is the indices of the initial seeds.

Geometrically, the initial seeds found by (\ref{eq:initial-seeds}) will have relatively high local densities
as well as high LDIs. Thus, they avoid being selected as outliers (due to the high local densities)
and avoid lying too close to each other (due to the high GDIs). As a result, these initial seeds can
lead to very fast convergence of the $k$-means algorithm when the clusters are separable.
This advantage will be verified by the experiments in Section \ref{sec:experiments}.

\subsection{Outliers detection and removal} \label{subsec:outliers}

Outliers detection and removal can be very useful for $k$-means to work stably.
% when their amount is large or effect is strong.
Here, we develop a simple algorithm to detect and remove the outliers, based on the local density and LDI.
First, we define the $\gamma^o$ value as follows:
\begin{equation}\label{eq:vo}
\gamma^o_i = \left(1-\frac{1}{2}\,\overline{\rho}_i^2 -\frac{1}{2}(1-\delta_i^l)^2\right)^2.
\end{equation}
This definition is very similar to that of $\gamma^c$ except that $\gamma^o$ is a decreasing function of $\rho$
and $\gamma^c$ is increasing with $\rho$ increasing. The points with low densities but high LDIs will get
high $\gamma^o$ values and are thus regarded as outliers.

Secondly, we use a threshold of $\gamma^o$, denoted by $\gamma^o_{t}$, to detect the outliers with the principle
that $\gamma^o$ of the outliers should be greater than $\gamma^o_{t}$. For example, if we set $\gamma^o_{t}=0.95$,
then the points with $\overline{\rho}<0.1$ and $\delta^l>0.8$ will have the $\gamma^o$ values being greater
than $0.95.$ Thus, they are treated as outliers. The set of outliers and the set of the corresponding indices
are denoted as $\bm{X}^o$ and $\bm{I}^o$, respectively. The other points with higher densities or lower LDIs
will get relatively smaller values of $\gamma^o$ and therefore will be treated as normal samples.

Finally, we remove the outliers from the data set $\bm{X}$ before $k$-means proceeds, that is,
setting $\bm{X}$$=$$\bm{X}\backslash\bm{X}^o$$=$$\{\bm{x}^{(i)}|1\leq i\leq m \text{~and~} i \notin\bm{I}^o\}$.

\subsection{Model selection for the LDPS algorithm}

\begin{figure}[htb]
\centering
\subfigure[$\bm{\theta}$-$k$ graph]{\includegraphics[width=0.46\linewidth]{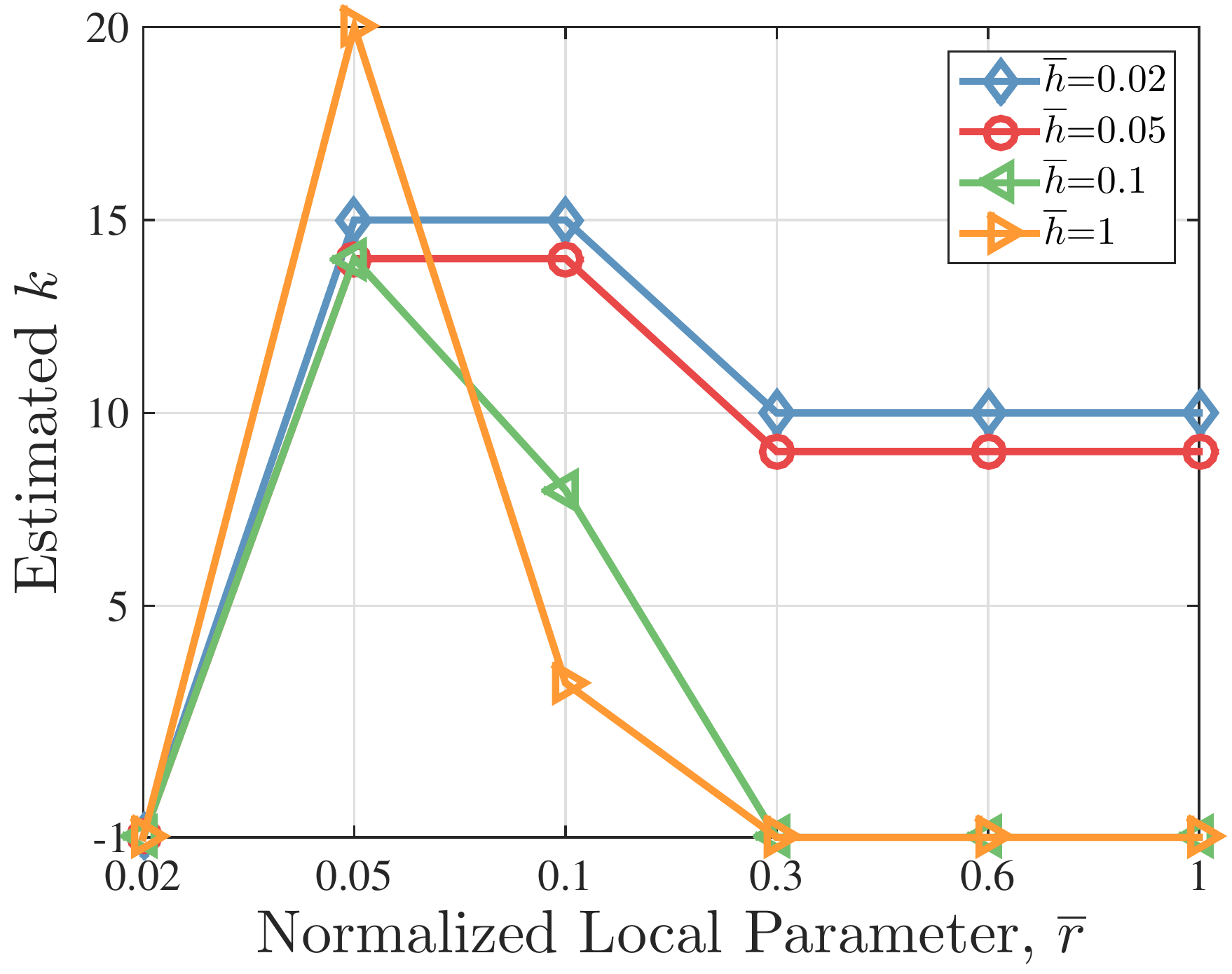}
\label{fig:R15-k}
}
\subfigure[$\bm{\theta}$-$\tau^*$ graph]{\includegraphics[width=0.46\linewidth]{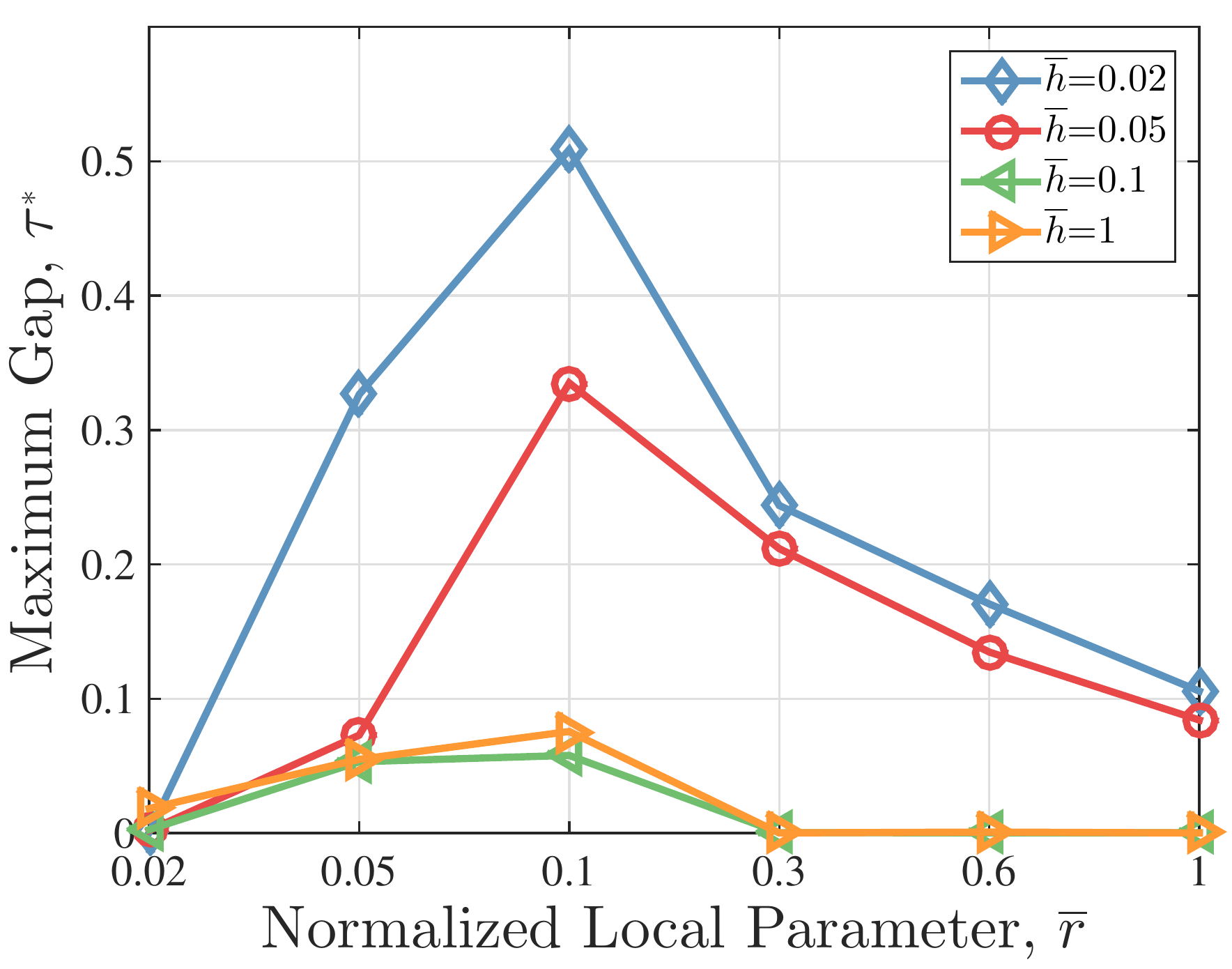}
\label{fig:R15-tau}
}
\caption{Effects of different parameters $\bm{\theta}$ for the LDPS algorithm on the R15 data set.
The ground truth is $15.$ (a) In the $\bm{\theta}$-$k$ graph, the estimated $k$ equals to $k^*$
when $\bm{\theta}=(0.02,0.05)$ or $\bm{\theta}=(0.02,0.01)$.  (b) In the $\bm{\theta}$-$\tau^*$ graph,
the maximum $\tau^*$ is achieved when $\bm{\theta}=(0.02,0.1)$, from which we get $k=k^*$.}
\label{fig:parameters-R15}
\end{figure}

The accuracy of the estimation of $k$ obtained by the LDPS algorithm depends heavily on the bandwidth $h$
for the local density estimation and the neighborhood size $r$ for the computation of LDI. Denote
by $\bm{\theta}=(\overline{h},\overline{r})$ the (normalized) parameters,
where $\overline{h}=h/d^*$ and $\overline{r}=r/d^*$. Fig. \ref{fig:R15-k} shows the results of the LDPS algorithm
with different parameters $\bm{\theta}$ on the R15 data set. As seen in Fig. \ref{fig:R15-k}, the estimated $k$
is equal to the ground truth $k^*$ only when the parameters $\bm{\theta}$ are properly selected.

\subsubsection{Parameters choosing by grid search}

In many real applications, we do not know beforehand what the true number $k^*$ of the clusters is.
Therefore, we need to define certain criteria to evaluate the estimated number $k$ of clusters and
do model selection to optimize the criteria.

Here, we utilize $\tau^*$ as a criterion to assess how good the estimated $k$ will be.
As discussed in Section \ref{subsec:estimate-k}, $\tau^*$ indicates the maximum gap of the $\gamma^c$ value
between the selected local density peaks and the other points. Mathematically, it can be written as a function
of the dissimilarity matrix $\bm{D}$ with parameters $\bm{\theta}$ (see Algorithm \ref{alg:ldps}).
The parameters that maximize $\tau^*$ will result in the most distinctive local density peaks. Thus,
we choose the parameters by solving the optimization problem:
\begin{equation}\label{eq:theta}
\hat{\bm{\theta}} = \argmax_{\bm{\theta}}~\tau^*(\bm{D}; \bm{\theta}).
\end{equation}
Fig. \ref{fig:R15-tau} shows the $\tau^*$ value on the R15 data set with respect to different $\bm{\theta}$,
where the dissimilarity measure is the square of the Euclidean distance.

There are no explicit solutions for the optimization problem (\ref{eq:theta}). A practical way of solving this problem
approximately will be the grid search method \cite{hsu2003practical}, in which various pairs of the $(\overline{h},\overline{r})$
values are tried and the one that results in the maximum $\tau^*$ is picked. Due to the local property of
the density estimation and LDI, $\overline{h}$ and $\overline{r}$ are generally set to lie in a small range
such as $(0,0.2]$ and $[0.05, 0.5]$, respectively. Take the R15 data set as an example, we equally split
$\overline{h}$ and $\overline{r}$ into $10$ fractions, respectively and then use a grid search procedure
to maximize $\tau^*$. The maximum gap we get is $\tau^*=0.53$ with $\hat{\bm{\theta}}=(0.02,0.1)$.

\iffalse
\begin{algorithm}[!thb]
\caption{Outliers Detection} \label{alg:outliers-detect}
\begin{algorithmic}[1]
\REQUIRE $\gamma^o_{thres}$
\STATE Initializing $\bm{I}^o$, $\bm{I}^o=\varnothing$;
\STATE Computing $\bm{\gamma}^o$ by (\ref{eq:vo});
\FOR{each $i\in [1,m]$}
\IF{$\bm{\gamma}^o_i>\gamma^o_{thres}$}
\STATE $\bm{I}^o=\bm{I}^o\cup \{i\}$\,;
\ENDIF
\ENDFOR
%\RETURN $\bm{I}_o$
\end{algorithmic}
\end{algorithm}
\fi

%=======================================================
%                   clustering
%=======================================================
\section{LDPS-means and LDPS-medoids} \label{sec:clustering}

In the previous section, we proposed the LDPS algorithm for initializing $k$-means and $k$-medoids.
For $k$-means, the input dissimilarity matrix $\bm{D}$ is the square of the Euclidean distance.
For $k$-medoids, any kind of dissimilarity matrix $\bm{D}$ can be used as input. In view of this difference,
they use different procedures for updating the cluster centers.

In this section, we propose two novel clustering algorithms, LDPS-means (Algorithm \ref{alg:ldps-means})
and LDPS-medoids (Algorithm \ref{alg:ldps-medoids}), as a combination of the LDPS initialization
algorithm (Algorithm \ref{alg:ldps}) with the clustering procedures of $k$-means and $k$-medoids, respectively.
Their clustering framework is implemented as in Fig. \ref{fig:framework}.

\begin{algorithm}[!thb]
\caption{The LDPS-means Algorithm}\label{alg:ldps-means}
\begin{algorithmic}[1]
\REQUIRE $\bm{X}$, $k^*$(optional), $h$, $r$, $\gamma^o_{t}$
\ENSURE $\bm{C}^*$, $\bm{X}^o$, and $\tau^*$
\STATE Perform the LDPS algorithm to get $\bm{\rho}$, $\bm{\delta^l}$, $\bm{I}_{ldp}$ and $\bm{\tau}$
\IF{$k^*$ is not given}
\STATE Estimate $k$ with $k=|\bm{I}_{ldp}|$, and set $k^*=k$
\ENDIF
\STATE Select $k^*$ initial seeds with indices $\bm{I}^s$ by (\ref{eq:initial-seeds});
\STATE Compute $\tau^*$:
$$\tau^* = \tau_{k^*}$$
\vspace{-0.4cm}
\STATE Detect outliers $\bm{X}^o$ with $\gamma^o_{t}$
\STATE Remove the outliers: $\bm{X}=\bm{X}\backslash \bm{X}^o$
\WHILE{not converging}
\STATE Compute assignments $\bm{S}$ by (\ref{eq:assignment})
\STATE Update the centers $\bm{C}$ by (\ref{eq:update-kmeans})
\ENDWHILE
\RETURN $\bm{C}^*=\bm{C}$, $\bm{X}^o$ and $\tau^*$
\end{algorithmic}
\end{algorithm}

\begin{algorithm}[!thb]
\caption{The LDPS-medoids Algorithm}
\label{alg:ldps-medoids}
\begin{algorithmic}[1]
\REQUIRE $\bm{D}$, $k^*$(optional), $h$, $r$, $\gamma^o_{t}$
\ENSURE $\bm{I}^{c*}$, $\bm{I}^o$ and $\tau^*$
\STATE Perform the LDPS algorithm to get $\bm{\rho}$, $\bm{\delta^l}$, $\bm{I}_{ldp}$ and $\bm{\tau}$
\IF{$k^*$ is not given}
\STATE Estimate $k$ with $k=|\bm{I}_{ldp}|$, and set $k^*=k$
\ENDIF
\STATE Select $k^*$ initial seeds with indices $\bm{I}^s$ by (\ref{eq:initial-seeds});
\STATE Compute $\tau^*$:
$$\tau^* = \tau_{k^*}$$
\vspace{-0.4cm}
\STATE Detect outliers with indexes $\bm{I}^o$
\STATE Remove the outliers:  $\bm{D}=(d_{ij})_{i,j\notin \bm{I}^{o}}$
\WHILE{not converging}
\STATE Compute assignments $\bm{S}$ by (\ref{eq:assignment})
\STATE Update the medoids by (\ref{eq:update-kmedoids})
\ENDWHILE
\RETURN $\bm{I}^{c*}=\bm{I}^c$, $\bm{I}^o$, and $\tau^*$
\end{algorithmic}
\end{algorithm}

LDPS-means is a powerful method to deal with spherically distributed data. However, it is unable to
separate non-spherically distributed clusters. LDPS-medoids can deal with this issue by choosing appropriate
dissimilarity measures. In the next subsection, we will discuss how to choose an appropriate dissimilarity measure
for the LDPS-medoids algorithm.

\subsection{Dissimilarity measures for LDPS-medoids}

A dissimilarity measure is the inverse of a similarity measure \cite{santini1999similarity}, which is a
real-word function that quantifies the similarity between two objects. It can be viewed as a kind of
distance without satisfying the distance axioms. It assesses the dissimilarity between data samples,
and the larger it is, the less similar the samples are.

Choosing appropriate dissimilarity measures for clustering methods is very crucial and task-specific.
The (square of) Euclidean distance is the most commonly used dissimilarity measure and suitable to deal
with spherically distributed data. The Mahalanobis distance is a generalization of the Euclidean distance and
can deal with hyper-ellipsoidal clusters. If the dissimilarity measure is the $L_1$ distance, $k$-medoids
will get the same result as $k$-median \cite{arya2004local}. For manifold distributed data,
the best choice for dissimilarity measures would be the manifold distance \cite{seung2000manifold} which is
usually approximated by the graph distance based on the $\epsilon$-neighborhood or the $t$-nearest-neighborhood ($t$-nn).
Graph-based $k$-means \cite{tu2014novel} uses this measure. For images, one of the most effective
similarity measures may be the complex wavelet structural similarity (CW-SSIM) index \cite{Sampat2009cwssim},
which is robust to small rotations and translations of images. In \cite{WCICA2016}, a combination of the manifold
assumption and the CW-SSIM index is used for constructing a new manifold distance named geometric CW-SSIM
distance, which shows a superior performance for visual object categorization tasks. Other cases include
the cosine similarity which is commonly used in information retrieval, defined on vectors arising from
the bag of words modal. In many machine learning applications, kernel functions such as the radial basis function (RBF)
kernel can be viewed as similarity functions.

In the section on experiments, whenever manifold-based dissimilarity measures are needed, we always use $t$-nn
as the neighborhood constructor for approximating the manifold distance. $t$ is generally set to be a small value such as $3,~5$ and $8$.

%=======================================================
%                   performance
%=======================================================
\section{Performance analysis} \label{sec:performance}

In this section, we analyze the performance of the local density and LDI based clustering methods.
To simplify the analysis, we make the following assumptions:
\begin{enumerate}[1)]
\item the clusters are spherical-distributed,
\item each cluster has a constant number ($m_0$) of data points,
\item the clusters are non-overlapping and can be separated by $k$-means with appropriate initial seeds.
\end{enumerate}
We use both the final SSE and the number of iterations (updates) as criteria to assess the performance of $k$-means
and LDPS-means. Under the above assumptions, we have the following results.

\begin{theorem}\label{the:performace}
Under Assumptions 1)-3) above, the average number of repeats that $k$-means needs to achieve the competitive
performance of LDPS-means is $O(e^k)$.
\end{theorem}

Here, \textit{competitive} means both good final SSE and less number of iterations. See Appendix % \ref{append1}
for the proof.

We now analyze the time complexity of $k$-means to achieve the competitive performance of LDPS-means.
The time complexity of LDPS-means is $O(m^2p+mkp)=O(m^2p)$. The time complexity of $k$-means to achieve
the competitive performance of LDPS-means is $O(E(\text{\#repeats})\cdot mkp)$.
This is summarized in the following theorem.

\begin{theorem}\label{the:time-complexity}
Under Assumptions 1)-3) above, the time complexity of $k$-means to achieve the competitive performance of
LDPS-means is $O(e^kmkp)$. The relative time complexity of $k$-means to achieve the competitive performance
of LDPS-means is $O(e^k/m_0)$.
\end{theorem}

Note that $k$ and $m_0$ do not depend on each other. Thus, compared with $k$-means, LDPS-means is superior
in time complexity when $k$ is much larger compared with $\log(m_0)$.

The above theorems are also true for $k$-medoids and LDPS-medoids, and in this case, the assumption 1) is not needed.

%=======================================================
%                   experiments
%=======================================================
\section{Experiments}\label{sec:experiments}

In this section, we conduct experiments to evaluate the effectiveness of the proposed methods.
The experiments mainly consist of two parts: one is for evaluating the performance of the LDPS algorithm
in estimating $k$ and the other is the clustering performances of LDPS-means and LDPS-medoids obtained by
the deterministic LDPS initialization algorithm (Algorithm \ref{alg:ldps}).
We also evaluate the effect of the outliers detection and removal procedure on the performance of
the clustering algorithm.

All experiments are conducted on a single PC with Intel i7-4770 CPU (4 Cores) and 16G RAM.

\subsection{The compared methods}

In the first part on estimating $k$, we compare the LDPS method with $x$-means, $dip$-means and CFSFDP on the estimation
of the cluster number $k$. The $x$-means is parameter-free. For $dip$-means, we set the significance level $\alpha=0$
for the dip test and the voting percentage $v_{thd}=1\%$ as in \cite{kalogeratos2012dip}. For CFSFDP, we follow the suggestion
in \cite{rodriguez2014clustering} to choose $d_c$ so that the average number of neighbors is around $1\%$ to $2\%$
of the total number of data points in the data set. Formula (\ref{eq:theta}) is used to estimate the parameters in
the LDPS algorithm. We denote LDPS with the square of the Euclidean distance and the manifold-based dissimilarity measure
as LDPS(E) and LDPS(M), respectively.

In the clustering part, we compare the clustering performance of LDPS-means and LDPS-medoids with $k$-means
and $k$-medoids, respectively.

\begin{table*}[!thb]
\centering
\caption{Results of the estimated $k$ on the synthetic data sets. (a) The A-sets have different numbers of clusters.
(b) The S-sets vary with the complexity of the data distribution. (c) The Dim-sets lie in the Euclidean space $\mathbb{R}^p$
with the varying dimensionality $p$. (d) The Shape-sets have different shapes.}\label{tab:estimate-k-synthetic}
\begin{small}
\begin{tabular}{|l|c!{\vrule width 0.8pt}c!{\vrule width 0.8pt}c!{\vrule width 0.8pt}c!{\vrule width 0.8pt}c!{\vrule width 0.8pt}c|}
\hline
\multicolumn{2}{|c!{\vrule width 0.8pt}}{Data Sets} & ~~$x$-means~~ & ~$dip$-means~ & ~~CFSFDP~~ & LDPS(E) & LDPS(M)  \\
\Xhline{0.8pt}
\multirow{4}{*}{(a). A-sets} & A${}_0(k^*=~5)$  & 1 & \textbf{5} & \textbf{5} & \textbf{5} & \textbf{5} \\ \cline{2-7}
 & A${}_1(k^*=20)$  & 2 & 14 & \textbf{20} & \textbf{20} & \textbf{20} \\ \cline{2-7}
 & A${}_2(k^*=35)$  & 1 & 30 & \textbf{35}  & \textbf{35} & \textbf{35} \\ \cline{2-7}
 & A${}_3(k^*=50)$  & 1 & 47 & 47 & \textbf{50}  & \textbf{50} \\ \Xhline{0.8pt}
\multirow{4}{*}{(b). S-sets} & S${}_1\,(\sigma=0.002, k^*=50)$ & 1 & 47 & \textbf{50} & \textbf{50} & \textbf{50} \\ \cline{2-7}
 & S${}_2\,(\sigma=0.004, k^*=50)$ & 1 & 43 & \textbf{50} & \textbf{50} & \textbf{50} \\ \cline{2-7}
 & S${}_3\,(\sigma=0.006, k^*=50)$ & 1 & 33 & 42 & \textbf{50} & \textbf{50} \\ \cline{2-7}
 & S${}_4\,(\sigma=0.008, k^*=50)$ & 1 & 30 & 35 & \textbf{50} & \textbf{50} \\ \Xhline{0.8pt}
\multirow{4}{*}{(c). Dim-sets} & D${}_3\,(p=3, k^*=50)$ & 1 & 11 & 41 & \textbf{50} & \textbf{50} \\ \cline{2-7}
 & D${}_{6}\,(p=6, k^*=50)$ & 1 & 1 & 37 & 47 & \textbf{50} \\ \cline{2-7}
 & D${}_{9}\,(p=9, k^*=50)$ & 1 & 1 & 30 & 48 & \textbf{50} \\ \cline{2-7}
 & D${}_{12}\,(p=12, k^*=50)$ & \textbf{50} & 1 & 25 & 47 & \textbf{50} \\ \Xhline{0.8pt}
\multirow{8}{*}{(d). Shape-sets} & Crescent ($k^*=2$) & 8 & \textbf{2} & 3 & 3 & \textbf{2} \\ \cline{2-7}
 & Flame ($k^*=2$) & 1  & 1 & 7 & 4 & \textbf{2}  \\ \cline{2-7}
 & Path-based ($k^*=3$) & 44 & 5 & 2 & 2 & \textbf{3}  \\ \cline{2-7}
 & Spiral ($k^*=3$) & 1 & 2 & \textbf{3} & \textbf{3} & \textbf{3}  \\ \cline{2-7}
 & Compound ($k^*=6$)  & 3  & 3 & 9 & 10 & \textbf{7}  \\ \cline{2-7}
 & Aggregation ($k^*=7$)  & 1  & 4 & 10 & 9 & \textbf{6}  \\ \cline{2-7}
 & R15 ($k^*=15$) & 1  & 8 & \textbf{15} & \textbf{15} & \textbf{15}  \\ \cline{2-7}
 & D31 ($k^*=31$) & 1  & 22 & \textbf{31} & \textbf{31} & \textbf{31} \\ \hline
\end{tabular}
\end{small}
\end{table*}

\subsection{The data sets}

\subsubsection{Overview of the data sets }

We use both synthetic data sets and real world data sets for evaluation. Four different kinds of synthetic data sets are
used: the A-sets \cite{A-sets} have different numbers of clusters, $k$, the S-sets \cite{S-sets} have different data
distributions, the Dim-sets vary with the dimensionality $p$ and the Shape-sets \cite{rodriguez2014clustering}
are of different shapes. They can be download from the clustering datasets
website \footnote{\url{http://cs.joensuu.fi/sipu/datasets/}}.
We made certain modifications on the S-sets and Dim-sets since these two sets are easy to be separated.
More details can be found in Section \ref{subsec:synthetic-data-sets}.
The real world data sets include Handwritten Pendigits \cite{alimoglu1996methods}, Coil-20 \cite{nene1996columbia},
Coil-100 \cite{nayar1996columbia} and Olivetti Face Database \cite{SamariaH94Olivetti}.

\subsubsection{Attribute normalization}

In clustering tasks, attribute normalization is an important preprocessing step to prevent the attributes with large
ranges from dominating the calculation of the distance. In addition, it helps to get more accurate numerical
computations. In our experiments, the attributes are generally normalized into the interval $[0,1]$
using the min-max normalization. Specifically,
$$
(\bm{x}_j^{(i)})_{\text{normalized}}= \frac{\bm{x}_j^{(i)}-\bm{x}_j^{min}}{\bm{x}_j^{max}-\bm{x}_j^{min}}
$$
where $\bm{x}_j^{(i)}$ is the $j$th attribute of the data point $\bm{x}^{(i)}$, $\bm{x}_j^{min}=\min_l\bm{x}_j^{(l)}$
and $\bm{x}_j^{max}=\max_l\bm{x}_j^{(l)}$. For gray images whose pixel values lying in $[0,255],$
we simply normalize the pixels by dividing $255.$

\subsection{Performance criteria} \label{sec:criteria}

For estimating $k$, we use the simple criterion that the better performance is achieved
when the estimated $k$ is close to the ground truth $k^*$.

For the task of clustering the synthetic data sets, three criteria are used to evaluate the performance
of initial seeds: 1) the total CPU time cost to achieve $\text{SSE}^*$, 2) the number of
repeats (\#repe) needed to achieve $\text{SSE}^*$ and 3) the number of assignment iterations (\#iter)
when $\text{SSE}^*$ is achieved in the repeat. We first run LDPS-means to get an upper bound
for $\text{SSE}^*$, denoted as $\text{SSE}^*_{0}$. We then run $k$-means repeatedly and
sequentially up to $1000$ times and record the minimal $\text{SSE}^*$, which is denote by $\text{SSE}^*_{k}$.
During this process, once $\text{SSE}^*_{k}$ is smaller than $\text{SSE}^*_{0}$, we record the current
number of repeats as \#repe and the number of iterations in this repeat as \#iter. Otherwise, \#repe
and \#iter are recorded when the minimal $\text{SSE}^*_{k}$ is achieved within the whole $1000$ repeats.
The same strategy is used to record the results of $k$-means++, whose minimal $\text{SSE}^*$ is denoted
as $\text{SSE}^*_{k++}$. Records of CPU time, \#iter, \#repe, $\text{SSE}^*_k$ and $\text{SSE}^*_{k++}$
are averaged over the $1000$ duplicate tests to reduce the randomness.

On the real world data sets, we consider the unsupervised classification task \cite{dueck2007non}.
Three criteria are used to evaluate the clustering performance of the comparing algorithms by comparing
the learned categories with the true categories. First, each learned category is associated with
the true category that accounts for the largest number of the training cases in the learned category
and thus the \textit{error rate} ($r_{e}$) can be computed. The second criterion is the
\textit{rate of true association} ($r_{t}$), which is the fraction of pairs of images from the same
true category that were correctly placed in the same learned category. The last criteria is
the \textit{rate of false association} ($r_{f}$), which is the fraction of pairs of images from
different true categories that were erroneously placed in the same learned category. The better
clustering performance is characterized with a lower value for $r_{e}$ and $r_{f}$ but a higher
value for $r_{t}$. To fairly compare the performance of LDPS-means with $k$-means, we make their
running time to be the same by control the number of repeats of $k$-means. The same strategy is applied
for LDPS-medoids and $k$-medoids. The results of $k$-means and $k$-medoids are recorded when $\text{SSE}^*$
in the repeat is the smallest one among the repeats.

\begin{table*}[!htb]
\caption{Clustering performance comparison on the A-sets with different numbers of clusters, where "time" stands for
the total CPU time cost in seconds, \#iter = the number of iterations when the optimal SSE is achieved,
\#repe = the number of repeats needed to achieve the optimal SSE and $\text{SSE}^*$ = the optimal SSE.
}\label{tab:clustering-synthetic-A}
\centering
\begin{tabular}{|c!{\vrule width 0.8pt}c|c|c|c!{\vrule width 0.8pt}c|c|c|c!{\vrule width 0.8pt}c|c|c|}
\hline
\multirow{2}{*}{Data Sets} & \multicolumn{4}{c!{\vrule width 0.8pt}}{$k$-means} & \multicolumn{4}{c!{\vrule width 0.8pt}}{$k$-means++}
& \multicolumn{3}{c|}{LDPS-means} \\ \cline{2-12}
& time (s) & {~}\#iter{~} & \#repe & $\text{SSE}^*_k$ & time (s) & {~}\#iter{~} & \#repe & $\text{SSE}^*_{k++}$
& time (s) & {~}\#iter{~} & $\text{SSE}^*_{0}$ \\ \Xhline{0.8pt}
A${}_0$ & \textbf{0.005} & 9.8  & 1.4 & \textbf{7.61} & 0.05 & 7.8  & 1.2    & \textbf{7.61}  & 0.17 & \textbf{4}
& \textbf{7.61} \\
\hline
A${}_1$ & \textbf{0.9}   & 15.7 & 76.2 & \textbf{6.75} & 1.48 & 14.4 & 23.9  & \textbf{6.75}  & 1.36 & \textbf{2}
& \textbf{6.75} \\
\hline
A${}_2$ & 101.6 & 15.6 & 4053 & 7.70 & 42.9  & 12.1  & 219  & \textbf{7.54}  & \textbf{3.49} & \textbf{2}
& \textbf{7.54} \\
\hline
A${}_3$ & 275.5 & 13.7 & 7411 & 7.78 & 2036  & 10   & 4612  & 7.08  & \textbf{6.94} &  \textbf{2}
& \textbf{6.99} \\
\hline
\end{tabular}
\end{table*}

\begin{table*}[!htb]
\caption{Clustering performances on S-sets and Dim sets with varying complexity of data distribution and with
varying dimensionality, respectively. The meanings of the criteria are the same as that in
TABLE \ref{tab:clustering-synthetic-A}}
\label{tab:clustering-synthetic-S-and-Dim}
\centering
\begin{tabular}{|c!{\vrule width 0.8pt}c|c|c|c!{\vrule width 0.8pt}c|c|c|c!{\vrule width 0.8pt}c|c|c|}
\hline
\multirow{2}{*}{Data Sets} & \multicolumn{4}{c!{\vrule width 0.8pt}}{$k$-means}
& \multicolumn{4}{c!{\vrule width 0.8pt}}{$k$-means++} & \multicolumn{3}{c|}{LDPS-means} \\
\cline{2-12}
& time (s) & {~}\#iter{~} & \#repe & $\text{SSE}^*_{k}$ & time (s) & {~}\#iter{~} & \#repe & $\text{SSE}^*_{k++}$
& time (s) & {~}\#iter{~} & $\text{SSE}^*_{0}$ \\
\Xhline{0.8pt}
S${}_1$ & 69.7  & 9.2  & 3972 & 3.21 & 728   & 4.6  & 2590  & \textbf{1.98}  & \textbf{3.09} & \textbf{2}
& \textbf{1.98} \\
\hline
S${}_2$ & 129   & 15   & 5486 & 4.61 & 1148  & 8.6  & 3967 & 4.03  & \textbf{3.39} & \textbf{2}
& \textbf{3.97}  \\
\hline
S${}_3$ & 142   & 16.7 & 5149 & 6.19 & 1027  & 11.8 & 3482 & 5.80  & \textbf{3.26} & \textbf{3}
& \textbf{5.73}  \\
\hline
S${}_4$ & 186   & 17.3 & 6038 & 7.25 & 654   & 16.5 & 2207 & \textbf{7.20}  & \textbf{3.29}
& \textbf{4}  & \textbf{7.20}  \\
\Xhline{0.8pt}
D${}_3$ & 132   & 11.7 & 5653 & 16.5 & 1675  & 7.2  & 5789 & 15.1  & \textbf{3.63} & \textbf{3}
& \textbf{14.6} \\
\hline
D${}_6$ & 111   & 16.3 & 3688 & 221  & 276   & 14   & 923  & \textbf{219}   & \textbf{3.15}
& \textbf{6}  & 221  \\
\hline
D${}_9$ & 140   & 13   & 5418 & 334  & 1245  & \textbf{11.3} & 4105 & 323   & \textbf{3.03} & 15
& \textbf{312} \\
\hline
D${}_{12}$ & 121 & 12.5 & 4566 & 650 & 291   & 10.2 & 953  & \textbf{630}   & \textbf{3.13}
& \textbf{7}  & 639 \\
\hline
\end{tabular}
\end{table*}

\begin{figure*}[!htb]
\centering
\subfigure[Crescent]{\includegraphics[width=0.22\linewidth]{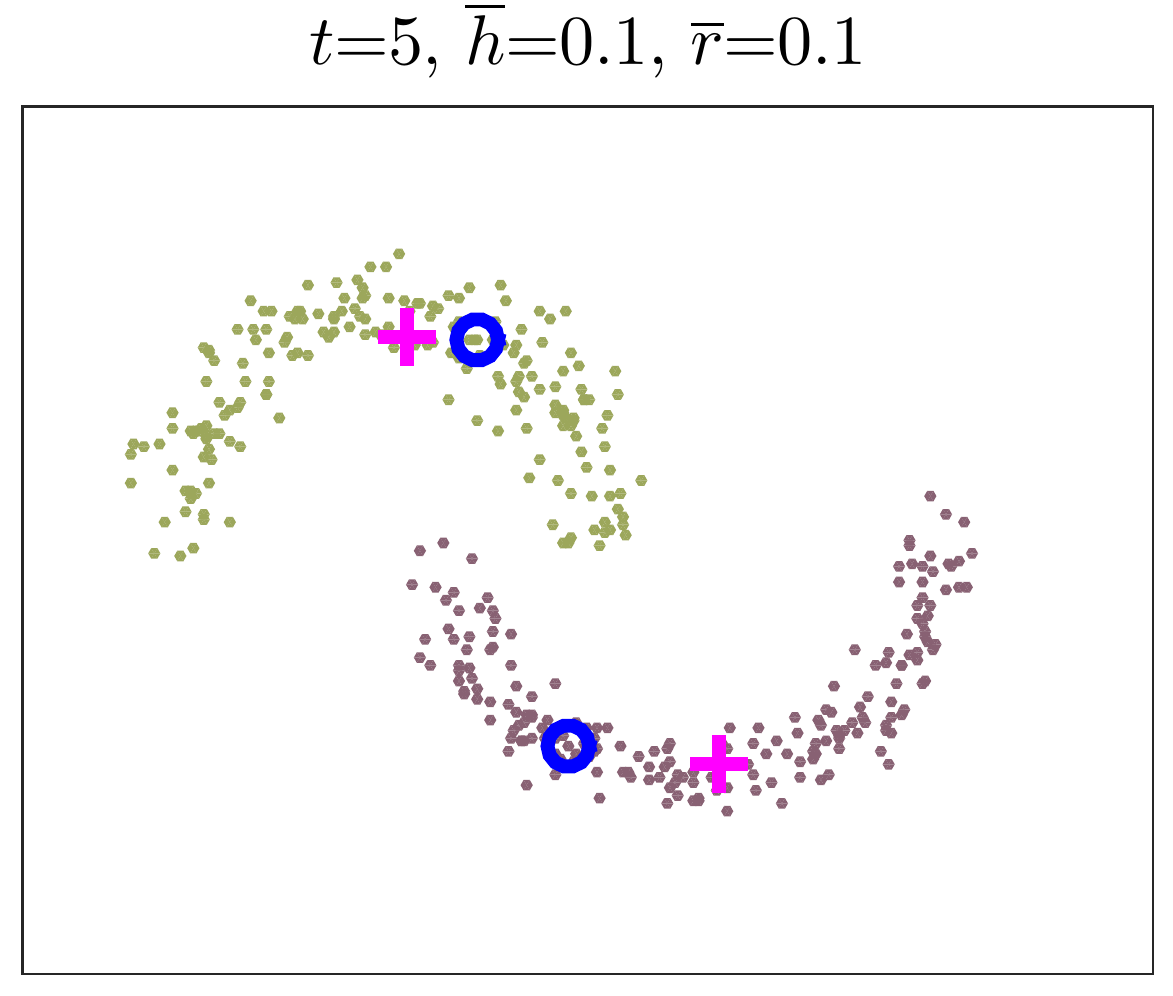}}
\text{\ }
\subfigure[Flame]{\includegraphics[width=0.22\linewidth]{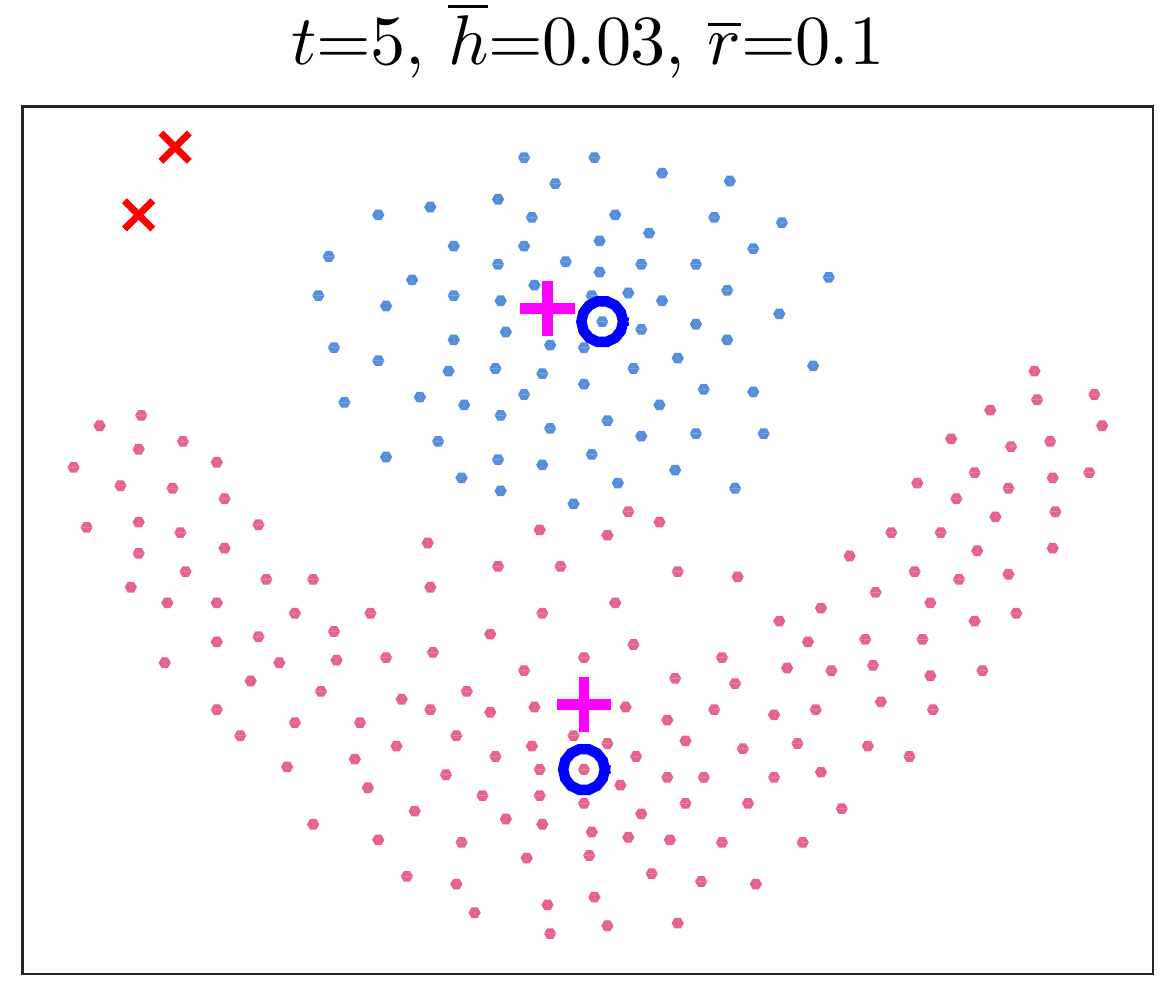}}
\text{\ }
\subfigure[Path-based]{\includegraphics[width=0.22\linewidth]{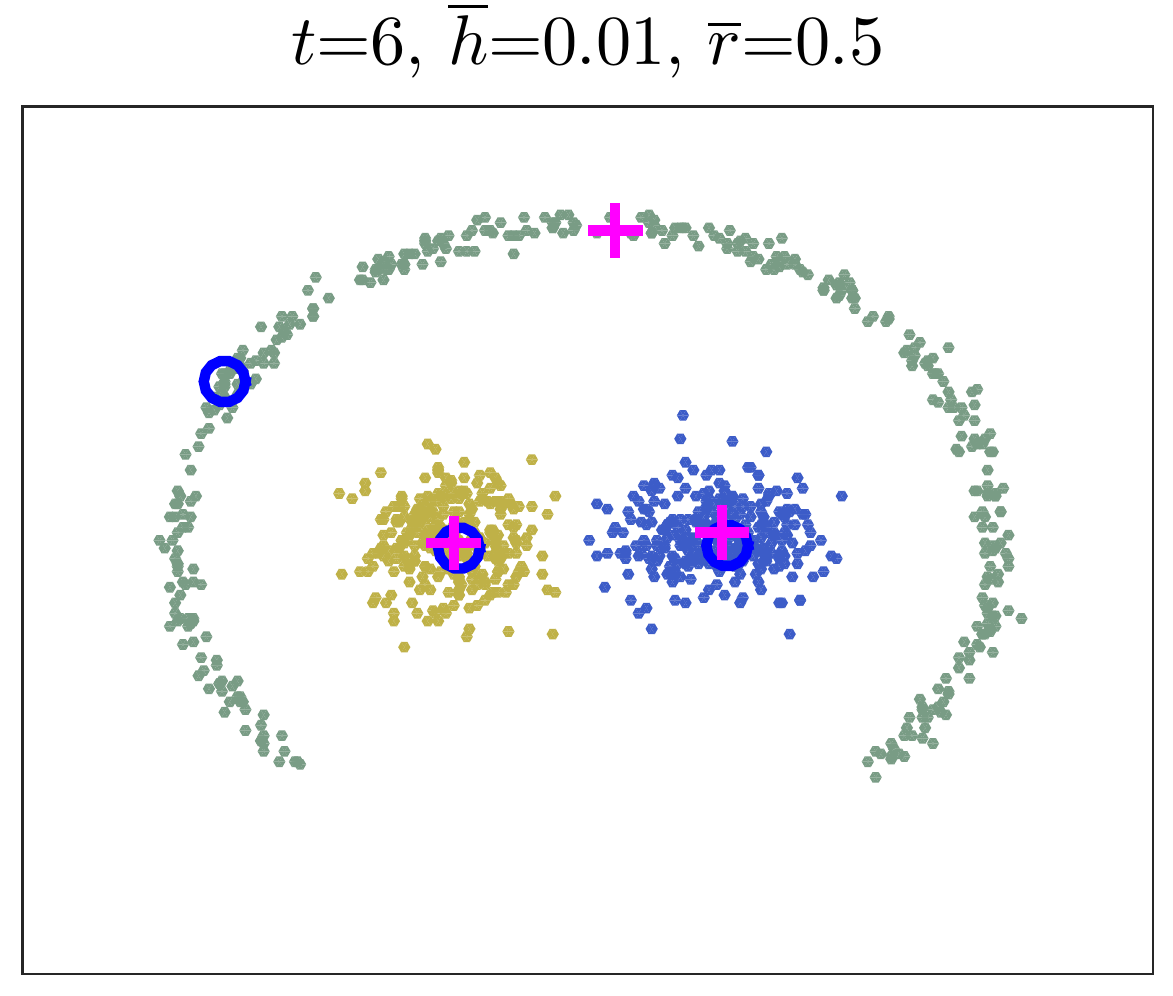}}
\text{\ }
\subfigure[Spiral]{\includegraphics[width=0.22\linewidth]{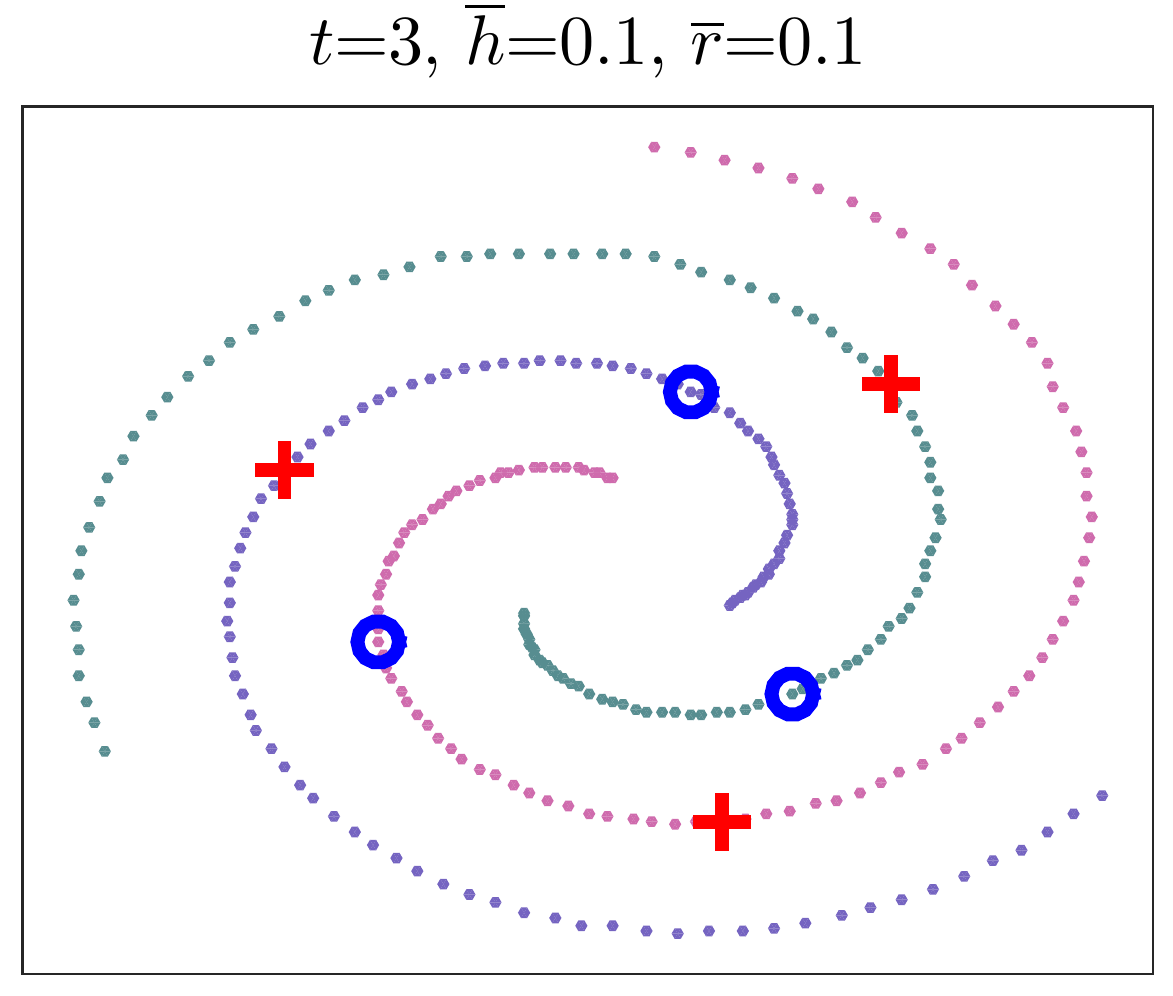}}
\subfigure[Compound]{\includegraphics[width=0.22\linewidth]{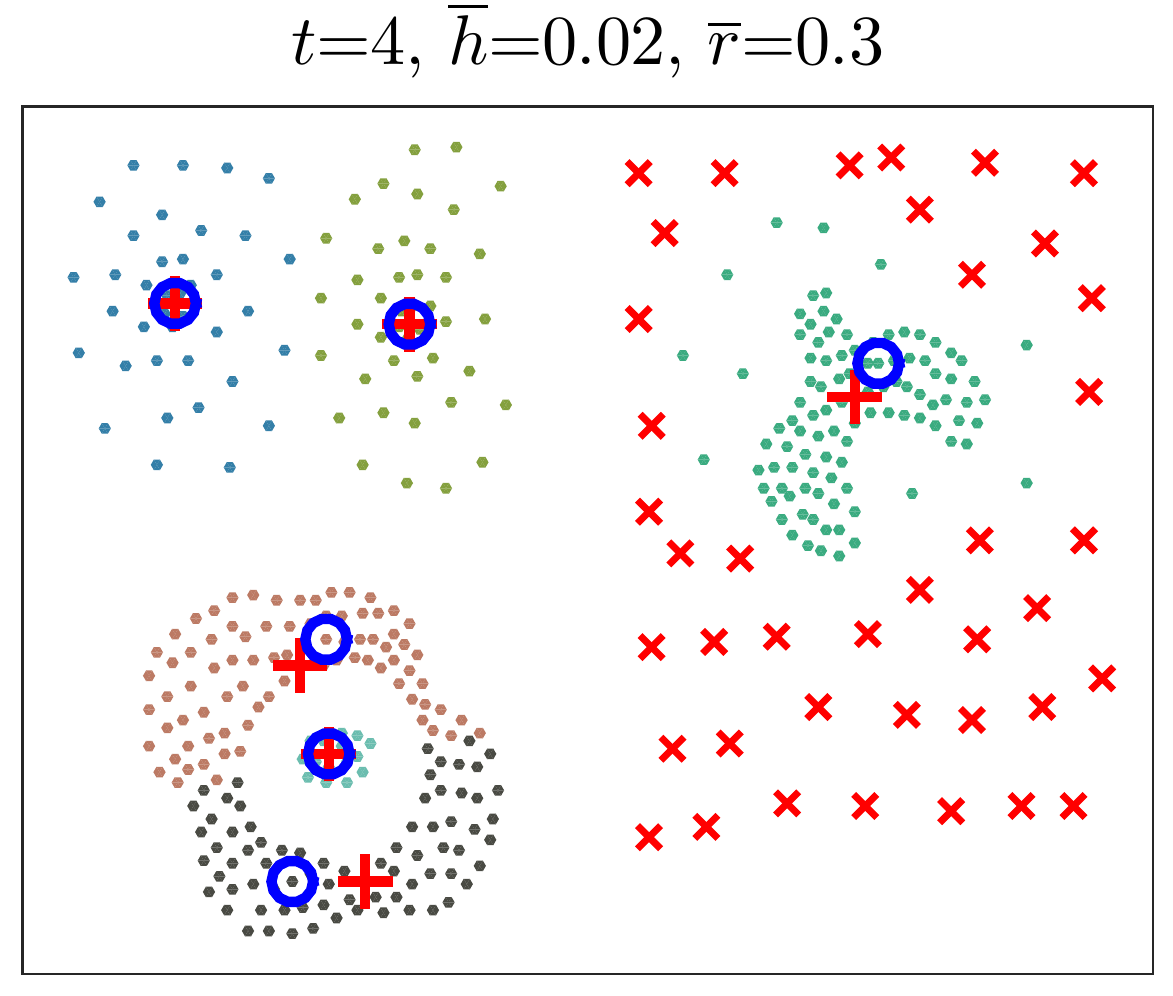}}
\text{\ }
\subfigure[Aggregation]{\includegraphics[width=0.22\linewidth]{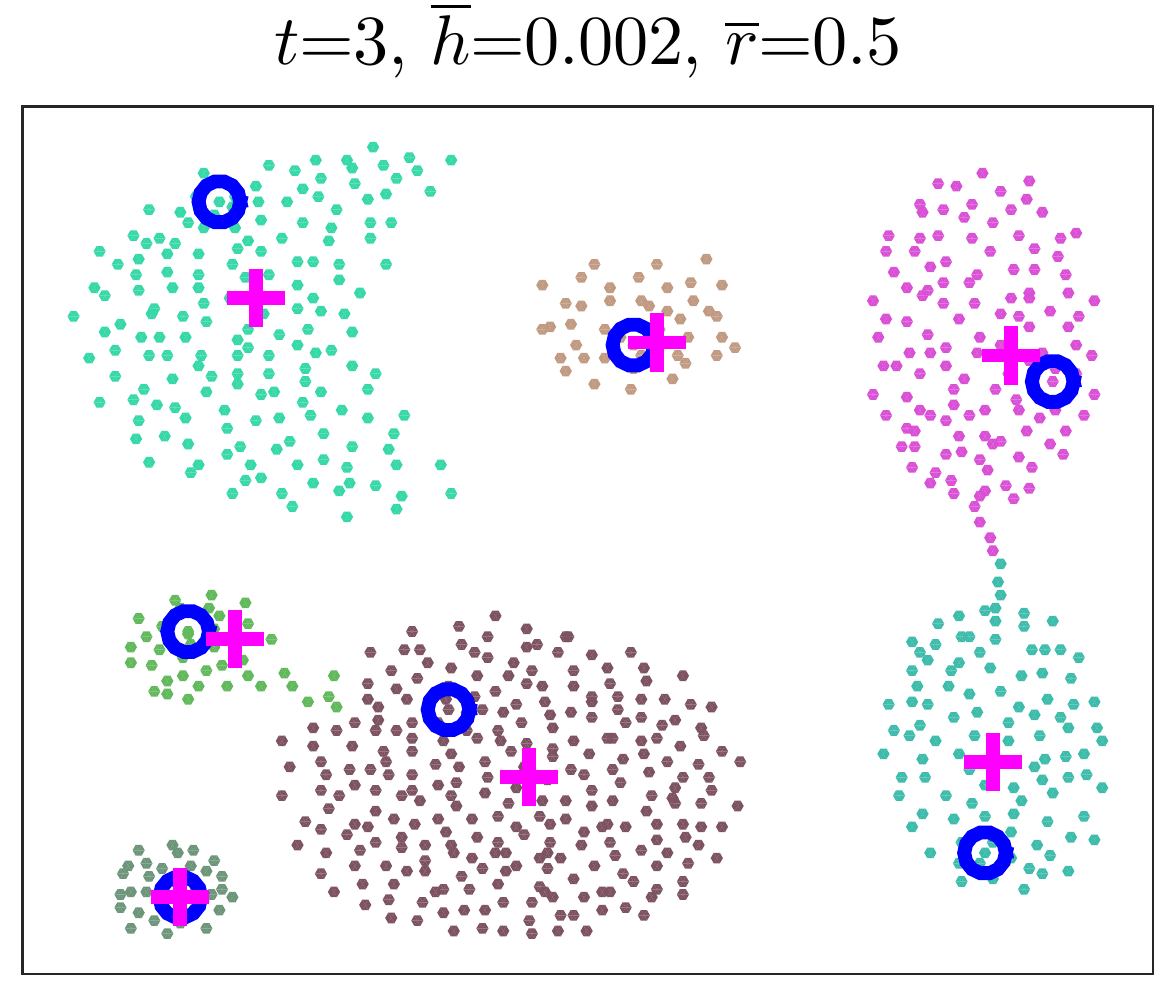}}
\text{\ }
\subfigure[R15]{\includegraphics[width=0.22\linewidth]{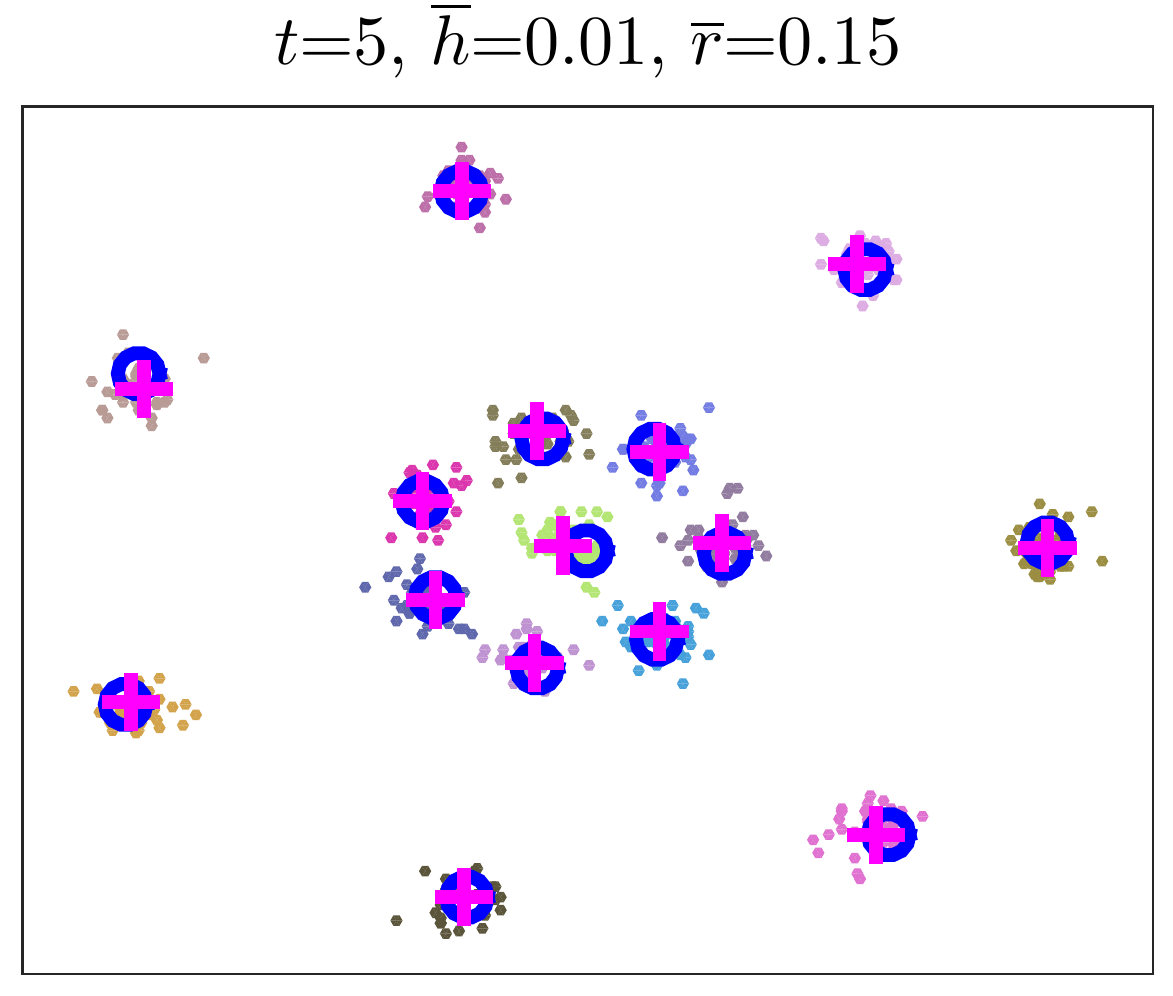}}
\text{\ }
\subfigure[D31]{\includegraphics[width=0.22\linewidth]{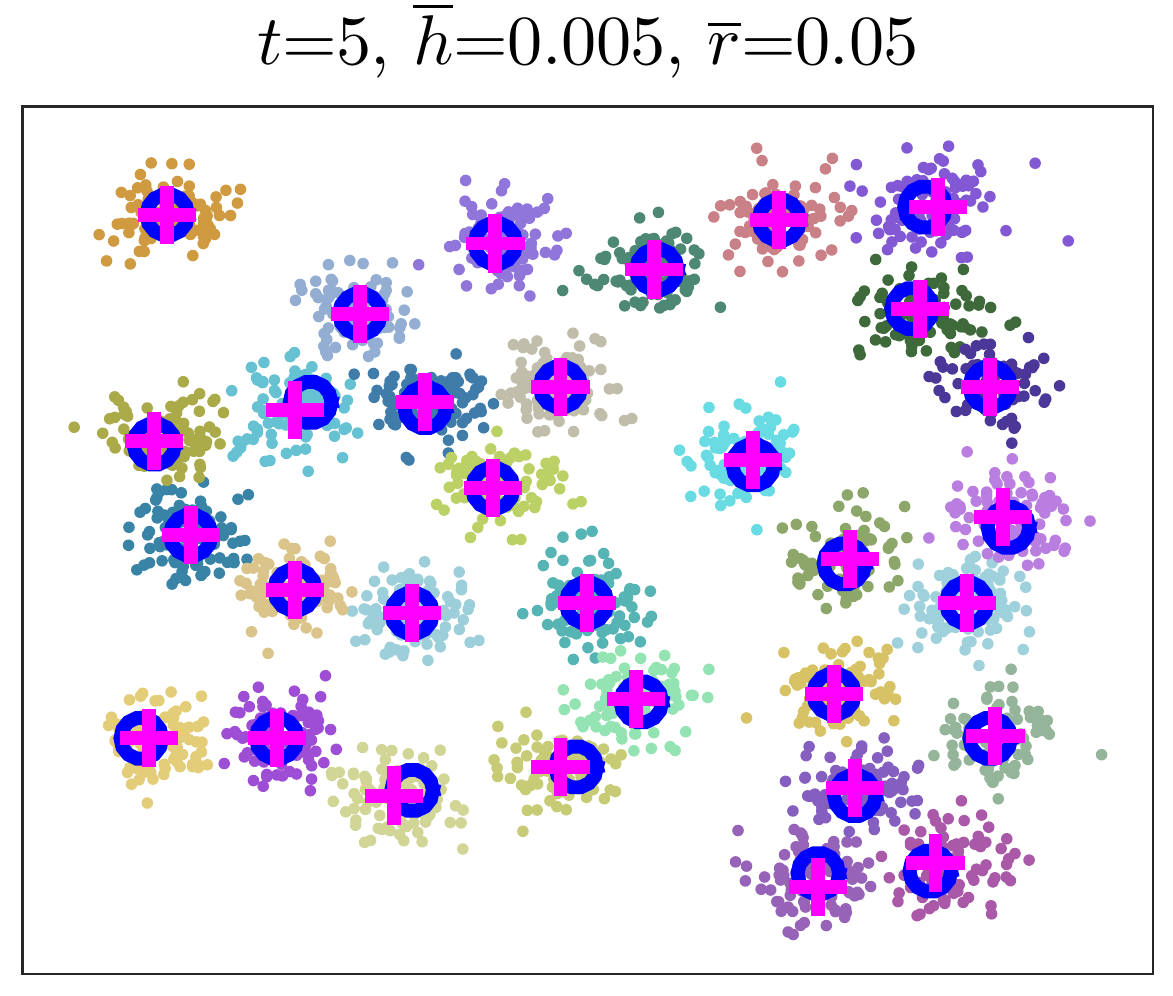}}
\caption{Visualization of the clustering results on the Shape-sets by LDPS-medoids.
Points marked with "$\bm{\circ}$" and "$\bm{+}$" are initial seeds and the final centroids, respectively.
Points marked with "$\bm{\times}$" are detected outliers. LDPS-medoids correctly detected the outliers in
the Flame set and most of the outliers in the Compound set.
}\label{fig:clustering-shapes}
\end{figure*}

\subsection{Experiments on synthetic data sets}\label{subsec:synthetic-data-sets}

We use four kinds of synthetic data sets. The A-sets contains three two-dimensional sets $A_1,$ $A_2$ and $A_3$
with different numbers of circular clusters ($k^*=20,35,50$). Each cluster has $150$ data points.
We generate a new set $A_0$ by selecting five clusters from $A_1$ with the labels $1-5.$

The S-sets $S_1$ to $S_4$ are composed of the data points sampled from two-dimensional Gaussian clusters
$N(\mathbf{\mu},\mathbf{\Sigma})$, with $100$ data points in each cluster. Their centers are the same as that of
the min-max normalized $A_3$ set. We set $\mathbf{\Sigma}=\sigma\cdot\mathbf{I}_2$, with $\sigma$
being $0.002, 0.004, 0.006$ and $0.008,$ respectively, where $\mathbf{I}_2$ is the identity matrix in $\mathbb{R}^2$.

We also generate four Dim-sets $D_p$ with the dimensionality $p= 3,6,9,12.$ The Dim-sets are
Gaussian clusters that distribute in multi-dimensional spaces. Each of them has $50$ clusters with $100$ data samples
in each cluster. The first two-dimensional projection of their cluster centers are the same as that of
the min-max normalized $A_3$ set. The axes in the other dimensions of the cluster centers are randomly distributed.
Their covariances are set to be $\sigma\cdot\mathbf{I}_p$ with $\sigma=0.001, 0.004, 0.007, 0.01,$
respectively, where $\mathbf{I}_p$ is the identity matrix in $\mathbb{R}^p$.

The Shape-sets consist of $8$ sets with different shapes (see Fig. \ref{fig:clustering-shapes}).
Six of them are the Flame set ($k^*$=2), the Spiral set ($k^*$=3), the Compound set ($k^*$=6),
the Aggregation set ($k^*$=7), the R15 set ($k^*$=15) and the D31 set ($k^*$=31).
We generate two new Shape-sets, the Crescent shape set ($k^*=2$) and the Path-based set ($k^*=3$).

\begin{table}[!bht]
\caption{Summary of the ability to learn $k$ with different data distributions, where \checkmark means "good",
$\circ$ means "general" and $\times$ means "bad".}\label{tab:clustering-synthetic-k}
\centering
\begin{tabular}{|c!{\vrule width 0.8pt}c|c|c|c|}
\hline
 & large $k$    & high $\sigma$ & large $p$      & shapes \\
\Xhline{0.8pt}
$x$-means    & $\times$   & $\times$      & $\circ$     & $\times$   \\
\hline
$dip$-means  & $\circ$    & $\circ$       & $\times$    & $\circ$   \\
\hline
CFSFDP       & \checkmark & $\circ$       & $\circ$     & $\circ$    \\
\hline
LDPS(E)      & \checkmark & \checkmark    & \checkmark  & $\circ$    \\
\hline
LDPS(M)      & \checkmark & \checkmark    & \checkmark  & \checkmark \\
\hline
\end{tabular}
\end{table}

\subsubsection{Performance on the estimation of $k$}

The results of the estimated $k$ are summarized in TABLE \ref{tab:estimate-k-synthetic}.
From this table it is seen that $x$-means fails to split on most of the data sets but it gets the correct result
for the $D_{12}$ set. $dip$-means gets better results than $x$-means in most of the cases, but it underestimated
$k$ for most of the data sets. In particular, $dip$-means fails to detect any valid cluster in $D_6$, $D_9$
and $D_{12}$ due to the relatively high dimensionality.
CFSFDP gets better results than $dip$-means on most of the data sets.
Though CFSFDP underestimated $k$ for the sets $A_3$, $S_3$, $S_4$ and all the Dim-sets,
it gets a very good estimation of $k$ (very close to the true cluster number $k^*$) on the Shape-sets.
Note that CFSFDP fails on the Flame set, the Compound set and the Aggregation set.
This is slightly different from the results reported in \cite{rodriguez2014clustering}.
It should be pointed out that the results reported in \cite{rodriguez2014clustering} can be achieved
with a very careful selection of parameters and with a prior knowledge on the data distribution,
which we did not consider in this paper.

\begin{table*}[!htb]
\caption{Results of the estimated $k$ on the Handwritten Pendigits data set and its subsets.}
\label{tab:estimate-k-pendigits}
\centering
%\begin{small}
\begin{tabular}{|c!{\vrule width 0.8pt}c!{\vrule width 0.8pt}c!{\vrule width 0.8pt}c!{\vrule width 0.8pt}
c!{\vrule width 0.8pt}c!{\vrule width 0.8pt}c!{\vrule width 0.8pt}c|}
\hline
~~Data sets~~ & ~~~$x$-means~~~ & ~~~$dip$-means~~~ & ~~~CFSFDP (E)~~~&~~~CFSFDP (M)~~~&~~LDPS (E)~~& ~LDPS (M) \\
\Xhline{0.8pt}
PD${}^{tr}_3(k^*=3)$ & 280  & 4 & \textbf{3} & 5 & \textbf{3} & \textbf{3} \\
\hline
PD${}^{tr}_5(k^*=5)$ & 453  & 2 & 4 & 4 & \textbf{5} & \textbf{5} \\
\hline
PD${}^{tr}_8(k^*=8)$ & 764  & 7 & 7 & 6 & \textbf{8} & \textbf{8} \\
\hline
PD${}^{tr}_{10}(k^*=10)$ & 942 & \textbf{9} & 7 & 8 & 8 & \textbf{11} \\
\Xhline{0.8pt}
PD${}^{te}_3(k^*=3)$ & 142 & 4 & 6 & \textbf{3} & \textbf{3} & \textbf{3} \\
\hline
PD${}^{te}_5(k^*=5)$ & 265 & 4 & \textbf{5} & \textbf{5} & \textbf{5} & \textbf{5} \\
\hline
PD${}^{te}_8(k^*=8)$ & 427 & 3 & 7 & 6 & 7 & \textbf{8} \\
\hline
PD${}^{te}_{10}(k^*=10)$ & 520 & 7 & 8 & 6 & 8 & \textbf{9} \\
\hline
\end{tabular}
\end{table*}

\begin{table*}[!htb]
\caption{Clustering performance comparison on the Handwritten Pendigits data set and its subsets,
where $r_{e}$ is the error rate, $r_{t}$ stands for the rate of true association which is the fraction of pairs
of images from the same true category that were correctly placed in the same learned category, and $r_{f}$ is the
rate of false association which is the fraction of pairs of images from different true categories that were
erroneously placed in the same learned category. }\label{tab:clustering-pendigits}
\centering
\begin{tabular}{|c!{\vrule width 0.8pt}c|c|c!{\vrule width 0.8pt}c|c|c!{\vrule width 0.8pt}c|c|c!{\vrule width 0.8pt}c|c|c|}
\hline
\multirow{2}{*}{Data Sets} & \multicolumn{3}{c!{\vrule width 0.8pt}}{$k$-means} & \multicolumn{3}{c!{\vrule width 0.8pt}}{LDPS-means}
& \multicolumn{3}{c!{\vrule width 0.8pt}}{$k$-medoids} & \multicolumn{3}{c|}{LDPS-medoids} \\ \cline{2-13}
& $r_{e}$ & $r_{t}$ & $r_{f}$ & $r_{e}$ & $r_{t}$ & $r_{f}$  & $r_{e}$  & $r_{t}$ & $r_{f}$ & $r_{e}$ & $r_{t}$ & $r_{f}$ \\
\Xhline{0.8pt}
PD${}^{tr}_3$  & 0.165 & 0.764 & 0.171 & 0.165 & 0.764 & 0.171 & \textbf{0.142} & \textbf{0.787} & \textbf{0.142}
& \textbf{0.142} & \textbf{0.787} & \textbf{0.142} \\
\hline
PD${}^{tr}_5$   & 0.220  & 0.826 & 0.111  & 0.079  & 0.856 & 0.037  & 0.054 & 0.906 & 0.029  & \textbf{0.046}
& \textbf{92.3} & \textbf{0.023} \\
\hline
PD${}^{tr}_8$  & 0.230  & 0.736 & 0.072  & 0.141  & 0.772 & 0.039  & 0.180 & 0.737 & 0.052  & \textbf{0.116}
& \textbf{81.6} & \textbf{0.031} \\
\hline
PD${}^{tr}_{10}$ & 0.290  & 0.664 & 0.066  & 0.256  & 0.641 & 0.051  & 0.207 & 0.757 & 0.041  & \textbf{0.157}
& \textbf{76.9} & \textbf{0.034} \\
\Xhline{0.8pt}
PD${}^{te}_3$   & 0.360  & \textbf{0.660} & 0.448 & 0.360 & \textbf{0.660} & 0.448 & \textbf{0.359} & \textbf{0.660}
& \textbf{0.446} & \textbf{0.359} & \textbf{0.660} & \textbf{0.446} \\
\hline
PD${}^{te}_5$    & 0.144 & 0.752 & 0.066  & 0.142 & 0.756 & 0.066  & \textbf{0.038}  & \textbf{0.934} & \textbf{0.019}
& \textbf{0.038}  & \textbf{0.934} & \textbf{0.019} \\
\hline
PD${}^{te}_8$    & 0.285 & 0.706 & 0.083  & 0.232 & 0.629 & 0.061  & 0.247 & 0.772 & 0.082  & \textbf{0.150}
& \textbf{0.806} & \textbf{0.042} \\
\hline
PD${}^{te}_{10}$ & 0.311 & 0.644 & 0.067  & 0.286 & 0.640 & 0.049  & 0.205 & 0.764 & 0.043  & \textbf{0.147}
& \textbf{0.801} & \textbf{0.031} \\
\hline
\end{tabular}
\end{table*}

LDPS(E) works very well on most of the data sets. Compared with CFSFDP, $lpds$-means obtained the correct $k$
on $A_3$, $S_3$ and $S_4$ and a very close $k$ to $k^*$ on the Dim-sets.
%This merit could be obtained only with the use of LDI rather than GDI.
Compared with the other comparing methods, LDPS(M) obtained the best results due to its use of LDI and
manifold-based dissimilarity measure. Compared with LDPS(E), LDPS(M) shows its superiority
in learning $k$ when dealing with the Shape-sets.

Based on the above analysis, we summarize the ability of the comparing methods for estimating $k$ on the synthetic data sets
in TABLE \ref{tab:clustering-synthetic-k}.

\subsubsection{Clustering performance}\label{subsec:performance-synthetic}

We first compare the clustering performance of LDPS-means with $k$-means and $k$-means++ on the A-sets to
verify the result of Theorem \ref{the:performace}. The experimental results are listed in
TABLE \ref{tab:clustering-synthetic-A}. As shown in TABLE \ref{tab:clustering-synthetic-A},
the clustering performance of LDPS-means is getting much better as $k$ increases.
On the sets $A_2$ and $A_3$, LDPS-means outperforms $k$-means and $k$-means++ greatly.
This is consistent with Theorem \ref{the:performace}.
%In addition, $k$-means can not achieve a competitive performance as LDPS-means on the $A_3$ set even with billions
%of repeats. This is consistent with Theorem \ref{the:performace}.

We then conduct experiments on S-sets and Dim-sets to show the capability of LDPS-means in separating clusters
with a varying complexity of data distributions and a varying dimensionality, respectively.
The experimental results are listed in TABLE \ref{tab:clustering-synthetic-S-and-Dim}.
From the table it is seen that, compared with $k$-means and $k$-means++, LDPS-means takes much less time and
much less number of iterations to achieve $\text{SSE}^*_{0}$. Note that $\text{SSE}^*_{0}$ is
smaller than $\text{SSE}^*_{k}$ and $\text{SSE}^*_{k++}$ on most of the data sets.

Finally, the variants of $k$-means (including $k$-means, $k$-means++ and LDPS-means) fail on most of the Shape-sets.
However, using LDPS-medoids with the manifold-based dissimilarity measure can get satisfactory results.
Fig. \ref{fig:clustering-shapes} shows the clustering results on the Shape-sets by LDPS-medoids
with an appropriate $t$ and the estimated parameters $\bm{\theta}=(\overline{h},\overline{r})$.

\subsection{Experiments on Handwritten Pendigits}

We now carry out experiments on the real world data set, Handwritten Pendigits, to evaluate the performance of
LDPS-means and LDPS-medoids on general purpose clustering. This data set can be download from the UCI Machine Learning repository\footnote{\url{http://archive.ics.uci.edu/ml/}}.
The Handwritten Pendigits data set contains totally $10992$ data points with $16$-dimensional features.
Each of them represents a digit from $0-9$ written by a human subject. The data set consists of a
training data set $PD^{tr}_{10}$ and a testing data set $PD^{te}_{10}$ with $7494$ and $3498$ samples,
respectively. Apart from the full data set, we also consider three subsets that contain the
digits \{1,3,5\} ($PD_3^{tr}$ and $PD_3^{te}$), \{0,2,4,6,7\} ($PD_5^{tr}$ and $PD_5^{te}$),
and \{0,1,2,3,4,5,6,7\} ($PD_8^{tr}$ and $PD_8^{te}$). On these sets, the manifold distance is
approximated by the graph distance, which is the shortest distance on the graph constructed by $5$-nn.

\subsubsection{Performance on the estimation of $k$}

TABLE \ref{tab:estimate-k-pendigits} presents the results of estimating $k$ on the Handwritten Pendigits.
$x$-means fails on all of those data sets. $dip$-means also fails on all of those data sets though it gets
the closest $k$ to the true cluster number on the $PD^{tr}_{10}$ set compared with all of the other comparing methods.
CFSFDP(E) and CFSFDP(M) get an underestimated $k$ at most of the cases. Compared with the CFSFDP methods,
the LDPS methods get the correct $k$ on most of the data sets owing to LDI. LDPS(M) gets better results than
those obtained by LDPS(E) on the $PD^{tr}_{10}$ and $PD^{te}_{10}$ sets due to the use of the manifold distance.

\subsubsection{Clustering performance}  \label{subsec:clustering-pendigits}

We now compare the unsupervised object classification performance of LDPS-mean and LDPS-medoids with $k$-means
and $k$-medoids. The results are shown in TABLE \ref{tab:clustering-pendigits}.
As seen in the table, LDPS-means gets better results than $k$-means does
on most of the data sets except for the $r_{true}$ criterion on $PD^{tr}_{10}$, $PD^{te}_{8}$ and $PD^{te}_{10}$,
where LDPS(E) fails to estimate the correct $k$. $k$-medoids gets better results than the other comparing methods
due to the use of the manifold distance as the dissimilarity measure.
However, LDPS-medoids gets the best results on all the data sets with all the criteria.

\begin{table*}[!htb]
\caption{Results of the estimated $k$ of the comparing methods on the Coil-sets.}
\label{tab:estimate-k-coil-sets}
\centering
\begin{tabular}{|c!{\vrule width 0.8pt}c!{\vrule width 0.8pt}c!{\vrule width 0.8pt}c!{\vrule width 0.8pt}c!
{\vrule width 0.8pt}c!{\vrule width 0.8pt}c|}
\hline
~~Data Sets~~ & ~~~$x$-means~~~ & ~~~$dip$-means~~~ & ~~~CFSFDP (E)~~~ & ~~~CFSFDP (M)~~~ & ~~LDPS (E)~~
& ~LDPS (M)~  \\
\Xhline{0.8pt}
Coil-5  & 74  & 7 & 2 & \textbf{5}  & 4  &  \textbf{5} \\
\hline
Coil-10 & 150 & 8 & 3 & 11 & 2  & \textbf{10} \\
\hline
Coil-15 & 215 & 7 & 4 & 14 & 6  & \textbf{15} \\
\hline
Coil-20 & 296 & 3 & 4 & 14 & 4  & \textbf{17} \\
\hline
\end{tabular}
\end{table*}

\begin{table*}[!htb]
\caption{Clustering performance comparison on the Coil-sets of the comparing methods,
where $r_{e}$ is the error rate, $r_{t}$ stands for the rate of true association which is the fraction of pairs
of images from the same true category that were correctly placed in the same learned category, and $r_{f}$ is the
rate of false association which is the fraction of pairs of images from different true categories that were
erroneously placed in the same learned category.
%Meanings of $r_{e}$, $r_{t}$ and $r_{f}$ are explained in Sec. \ref{sec:criteria}.
}\label{tab:clustering-coil-sets}
\centering
\begin{tabular}{|c!{\vrule width 0.8pt}c|c|c!{\vrule width 0.8pt}c|c|c!{\vrule width 0.8pt}c|c|c!{\vrule width 0.8pt}c|c|c|}
\hline
\multirow{2}{*}{Data Sets} & \multicolumn{3}{c!{\vrule width 0.8pt}}{$k$-means}
& \multicolumn{3}{c!{\vrule width 0.8pt}}{LDPS-means} & \multicolumn{3}{c!{\vrule width 0.8pt}}{$k$-medoids}
& \multicolumn{3}{c|}{LDPS-medoids} \\
\cline{2-13}
& $r_{e}$ & $r_{t}$ & $r_{f}$ & $r_{e}$ & $r_{t}$ & $r_{f}$  & $r_{e}$  & $r_{t}$ & $r_{f}$ & $r_{e}$ & $r_{t}$ & $r_{f}$ \\
\Xhline{0.8pt}
Coil-5  & 0.317 & 0.582 & 0.135  & 0.478 & 0.426 & 0.158 & \textbf{0.025} & \textbf{0.956} & \textbf{0.013}  & \textbf{0.025}
& \textbf{0.956} & \textbf{0.013}   \\
\hline
Coil-10 & 0.169 & 0.823 & 0.039  & 0.226 & 0.809 & 0.050  & 0.114 & 0.932 & 0.028  & \textbf{0.030} & \textbf{0.957}
& \textbf{0.007}  \\
\hline
Coil-15 & 0.246 & 0.778 & 0.027  & 0.237 & 0.786 & 0.032  & 0.221 & 0.879 & 0.064  & \textbf{0.029} & \textbf{0.956}
& \textbf{0.004} \\
\hline
Coil-20 & 0.336 & 0.664 & 0.028  & 0.353 & 0.609 & 0.028  & 0.281 & 0.821 & 0.039  & \textbf{0.171} & \textbf{0.846}
& \textbf{0.016}  \\
\hline
\end{tabular}
\end{table*}

\begin{table*}[!htb]
\caption{Results of the estimated $k$ of the comparing methods on the large Coil-sets.}
\label{tab:estimate-k-coil-sets-large}
\centering
\begin{tabular}{|c!{\vrule width 0.8pt}c!{\vrule width 0.8pt}c!{\vrule width 0.8pt}c!{\vrule width 0.8pt}c!{\vrule width 0.8pt}c!{\vrule width 0.8pt}c|}
\hline
~~Data Sets~~ & ~~~$x$-means~~~ & ~~~$dip$-means~~~ & ~~~CFSFDP (E)~~~ & ~~~CFSFDP (M)~~~ & ~~LDPS (E)~~ & ~LDPS (M)~  \\ \Xhline{0.8pt}
Coil-25  & 369  & 8  & 5   & 14  & 5  &  \textbf{24} \\ \hline
Coil-50  & 691  & 7  & 13  & 36 & 7  &  \textbf{49} \\ \hline
Coil-75  & 1069 & 15 & 18  & 45 & 4  &  \textbf{74} \\ \hline
Coil-100 & 1349 & 20 & 24 & 56 & 19  & \textbf{97} \\ \hline
\end{tabular}
\end{table*}

\begin{table*}[!htb]
\caption{Clustering performance comparison on the large Coil-sets,
where $r_{e}$ is the error rate, $r_{t}$ stands for the rate of true association which is the fraction of pairs
of images from the same true category that were correctly placed in the same learned category, and $r_{f}$ is the
rate of false association which is the fraction of pairs of images from different true categories that were
erroneously placed in the same learned category.
%Meanings of $r_{e}$, $r_{t}$ and $r_{f}$ are explained in Sec. \ref{sec:criteria}.
}\label{tab:clustering-coil-sets-large}
\centering
\begin{tabular}{|c!{\vrule width 0.8pt}c|c|c!{\vrule width 0.8pt}c|c|c!{\vrule width 0.8pt}c|c|c!{\vrule width 0.8pt}c|c|c|}
\hline
\multirow{2}{*}{Data Sets} & \multicolumn{3}{c!{\vrule width 0.8pt}}{$k$-means}
& \multicolumn{3}{c!{\vrule width 0.8pt}}{LDPS-means} & \multicolumn{3}{c!{\vrule width 0.8pt}}{$k$-medoids}
& \multicolumn{3}{c|}{LDPS-medoids} \\
\cline{2-13}
& $r_{e}$ & $r_{t}$ & $r_{f}$ & $r_{e}$ & $r_{t}$ & $r_{f}$  & $r_{e}$  & $r_{t}$ & $r_{f}$ & $r_{e}$ & $r_{t}$ & $r_{f}$ \\
\Xhline{0.8pt}
Coil-25  & 0.323 & 0.674 & 0.023 & 0.372 & 0.665 & 0.031 & 0.226 & 0.801 & 0.022 & \textbf{0.180} & \textbf{0.843}
& \textbf{0.016}   \\
\hline
Coil-50  & 0.371 & 0.624 & 0.016 & 0.406 & 0.643 & 0.016 & 0.325 & 0.741 & 0.024 & \textbf{0.199} & \textbf{0.829}
& \textbf{0.010}  \\
\hline
Coil-75  & 0.429 & 0.572 & 0.010 & 0.484 & 0.581 & 0.013 & 0.341 & 0.698 & 0.017 & \textbf{0.239} & \textbf{0.764}
& \textbf{0.007}   \\
\hline
Coil-100 & 0.437 & 0.551 & 0.009 & 0.472 & 0.507 & \textbf{0.008} & 0.415 & 0.657 & 0.019 & \textbf{0.271}
& \textbf{0.749} & 0.010  \\
\hline
\end{tabular}
\end{table*}

\subsection{Experiments on Coil-20 }

\begin{figure}[!bht]
\centering
\includegraphics[width=0.9\linewidth]{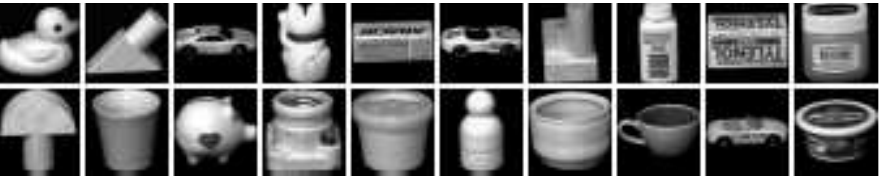}
\caption{Example objects of Coil-20. Their labels are the same as their order, sorting from left to right
and then up to down.}\label{fig:coil-20-examples}
\end{figure}

We now consider the real world data set, Coil-20 \cite{nene1996columbia}, which is used for the task of unsupervised
object clustering.
The data set Coil-20 contains $20$ objects, and each object contains $72$ images taken $5$ degree apart as
the object rotated on a turntable. We compress each image into $32\times32$ pixels with $256$ grey levels per pixel.
To compare the performance of the comparing methods with different numbers of categories, we select three subsets of
Coil-20: Coil-5 (objects 1,3,5,7 and 9), Coil-10 (objects with even number), Coil-15 (objects except 3,7,11,15 and 19).
Fig. \ref{fig:coil-20-examples} shows some examples of the twenty objects.

\text{~}

We use the CW-SSIM index and $t$-nn to construct the manifold distance. First, the CW-SSIM index is performed
on the images to get the structural similarity between the objects. Then, we construct the graph distance
using the $3$-nn neighborhood based on the structural similarity. Finally, the graph distance is served as
an approximation to the manifold distance between the objects.

\subsubsection{Performance on the estimation of $k$}

The estimated cluster number $k$ of Coil-20 and its subsets is presented in TABLE \ref{tab:estimate-k-coil-sets}.
$x$-means again fails to get a reasonable estimation of $k$ for these sets. $dip$-means, CFSFDP(E) and LDPS(E) also
fails to get a meaningful number of clusters since the Euclidean distance can not properly measure the
dissimilarity between these objects. Compared with these four method, CFSFDP(M) gets betters results though it
underestimated $k$ on Coil-15 and Coil-20. LDPS(M) obtained reasonable estimations of $k$ for most of the data sets.
It underestimates $k$ on Coil-20 since the objects 3, 6 and 19 are very similar (see Fig. \ref{fig:coil-20-examples});
they are clustered to the same category "cars" by the LDPS(M) algorithm.
Similarly, the objects 15 and 17 have very similar shapes and are thus clustered to the same category.

\subsubsection{Clustering performance}

TABLE \ref{tab:clustering-coil-sets} shows the clustering results on the Coil-sets. Unlike the results on the Pendigits sets,
LDPS-means has got a worse result than $k$-means did on these sets. This may be because LDPS(E) failed to estimate
the proper $k$ on the Coil-sets. LDPS-medoids, on the contrary, learnt the proper $k$ on these sets and selected the initial
seeds with high quality. As a result, the clustering performance of LDPS-medoids is much better than that of all the other
comparing methods on the Coil-sets.

\subsection{Experiments on Coil-100}

Coil-100 is the third real world data set we considered. Unlike Coil-20 which has a small number of categories,
the true cluster number $k^*$ of Coil-100 is very large. Thus, it is used for the task of unsupervised object
clustering with a large number of categories. The Coil-100 data set contains $100$ categories of objects consisting
of $7200$ color images, and each object has $72$ images taken $5$ degree apart as the object rotated on a turntable.
The color images are converted into gray images and resized to $32\times32$ pixels. We select three subsets from
Coil-100, which are Coil-25 (objects 1, 5, 9, $\cdots$, 93 and 97), Coil-50 (objects with even number)
and Coil-75 (Coil-25 + Coil-50).

The manifold distance is approximated using the graph distance with the CW-SSIM index and the $4$-nn neighborhood.

\begin{table*}[!htb]
\caption{Results of the estimated $k$ of the comparing methods on the Oliv.-sets.}
\label{tab:estimate-k-olivettifaces}
\centering
\begin{tabular}{|c!{\vrule width 0.8pt}c!{\vrule width 0.8pt}c!{\vrule width 0.8pt}c!{\vrule width 0.8pt}c!
{\vrule width 0.8pt}c!{\vrule width 0.8pt}c|}
\hline
~~Data Sets~~ & ~~~$x$-means~~~ & ~~~$dip$-means~~~ & ~~~CFSFDP (E)~~~ & ~~~CFSFDP (M)~~~ & ~~LDPS (E)~~ & ~LDPS (M)~ \\
\Xhline{0.8pt}
Oliv.-10 & 18 & 1 & 4 & \textbf{10} & 4  & \textbf{10} \\
\hline
Oliv.-20 & 33 & 1 & 1 & 18 & 5  & \textbf{20} \\
\hline
Oliv.-30 & 45 & 1 & 1 & 21 & $-1$  & \textbf{30} \\
\hline
Oliv.-40 & 56 & 1 & 1 & 28 & $-1$  & \textbf{36} \\
\hline
\end{tabular}
\end{table*}

\begin{table*}[!htb]
\caption{Clustering performance comparison of the comparing methods on the Oliv.-sets,
where $r_{e}$ is the error rate, $r_{t}$ stands for the rate of true association which is the fraction of pairs
of images from the same true category that were correctly placed in the same learned category, and $r_{f}$ is the
rate of false association which is the fraction of pairs of images from different true categories that were
erroneously placed in the same learned category.
%Meanings of $r_{e}$, $r_{t}$ and $r_{f}$ are explained in Sec. \ref{sec:criteria}.
}\label{tab:clustering-olivettifaces}
\centering
\begin{tabular}{|c!{\vrule width 0.8pt}c|c|c!{\vrule width 0.8pt}c|c|c!{\vrule width 0.8pt}c|c|c!{\vrule width 0.8pt}c|c|c|}
\hline
\multirow{2}{*}{Data Sets} & \multicolumn{3}{c!{\vrule width 0.8pt}}{$k$-means} &\multicolumn{3}{c!{\vrule width 0.8pt}}{LDPS-means}
& \multicolumn{3}{c!{\vrule width 0.8pt}}{$k$-medoids} & \multicolumn{3}{c|}{LDPS-medoids} \\
\cline{2-13}
& $r_{e}$ & $r_{t}$ & $r_{f}$ & $r_{e}$ & $r_{t}$ & $r_{f}$  & $r_{e}$  & $r_{t}$ & $r_{f}$ & $r_{e}$ & $r_{t}$ & $r_{f}$ \\
\Xhline{0.8pt}
Oliv.-10 & 0.240 & 0.804 & 0.052 & 0.200 & 0.849 & 0.041 & 0.110 & 0.896 & 0.022 & \textbf{0.070}  & \textbf{0.907}
& \textbf{0.016} \\
\hline
Oliv.-20 & 0.415 & 0.492 & 0.042 & 0.395 & 0.511 & 0.043 & 0.250 & 0.729 & \textbf{0.031} & \textbf{0.245} & \textbf{0.740}
& 0.032 \\
\hline
Oliv.-30 & 0.427 & 0.458 & 0.027 & 0.403 & 0.529 & 0.025 & 0.280 & 0.717 & 0.020 & \textbf{0.197} & \textbf{0.764}
& \textbf{0.013} \\
\hline
Oliv.-40 & 0.455 & 0.465 & 0.026 & 0.465 & 0.482 & 0.025 & 0.275 & 0.688 & 0.021 & \textbf{0.213} & \textbf{0.739}
& \textbf{0.010} \\
\hline
\end{tabular}
\end{table*}

\subsubsection{Performance on the estimation of $k$}

Since the number of clusters of Coil-100 is large, the density of certain local density peaks can be easily dominated by
the largest density. Thus, it is very hard to get a balanced local density distribution for the local density
peaks. To deal with this difficulty the local density is normalized as follows:
\be\label{eq:rho2}
\overline{\rho}=(\frac{\rho}{\rho^*})^{{1}/{4}}.
\en
The main purpose of the local density normalization (\ref{eq:rho2}) is to enlarge the relatively small local densities.
In this case, $\overline{\rho}$ also lies in the range of $(0,1]$ and is in the same increasing order as the original $\rho$.

TABLE \ref{tab:estimate-k-coil-sets-large} shows the results of the estimated cluster number $k$ on the large Coil-sets.
The LDPS methods use the new normalized local density $\overline{\rho}$ (defined in (\ref{eq:rho2}))
as the local density value. Surprisingly, LDPS(M) has learnt almost the identical number of clusters as $k^*$
with the new normalized local density $\overline{\rho}$ and the manifold distance. CFSFDP(M),
however, underestimated the cluster number $k$ on all of the data sets.
The other comparing methods all fail to learn a reasonable cluster number $k$.

\begin{figure}[!htb]
\centering
\includegraphics[width=0.6\linewidth]{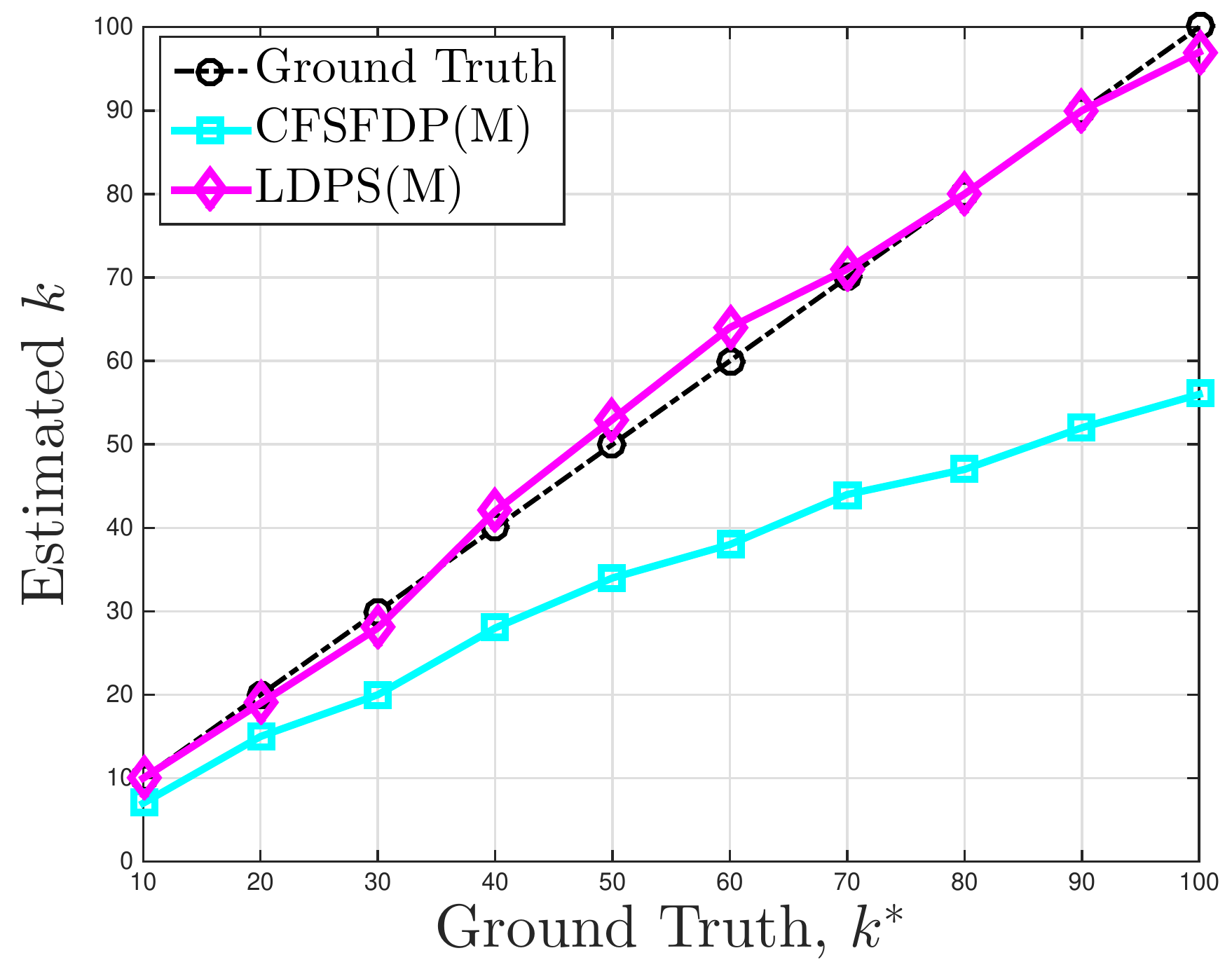}\\
\caption{The $k$-$k^*$ curves of LDPS(M) and CFSFDP(M) on the subsets of Coil-100. The horizontal axis is
the ground truth of the number of clusters and the vertical axis is the learned $k$.
}\label{fig:estimate_k_coil_100_CFSFDP_LDPS}
\end{figure}

To obtain a more careful comparison of the ability to learn $k$ with LDPS(M) and CFSFDP(M) on Coil-100,
we select the first $k^*$ categories ($k^*=10,20,\cdots,100$) from Coil-100 as $k^*$ subsets.
The $k^*$-$k$ curves are shown in Fig. \ref{fig:estimate_k_coil_100_CFSFDP_LDPS}. As $k^*$ is getting bigger,
the estimated $k$ by CFSFDP(M) is getting farther away from $k^*$; however, the $k$ learned
by LDPS(M) is getting much closer to the ground truth $k^*$ with very little changes. Thus, our method is much more
effective in learning $k$ than CFSFDP when the ground truth $k^*$ is very large.
This is consist with the analysis in Section \ref{subsec:dens-and-ldi}.

\subsubsection{Clustering performance}

The new normalized local density $\overline{\rho}$, defined in (\ref{eq:rho2}), is also used for this task.
Based on new normalized local density $\overline{\rho}$, the clustering results are shown in
TABLE \ref{tab:clustering-coil-sets-large}. LDPS-medoids again gets the best clustering results compared with
the other comparing methods. In addition, with $k^*$ increasing, the relative performance of LDPS-means is getting better
compared with $k$-medoids. This is consist with the conclusion in Theorem \ref{the:time-complexity}.

\subsection{Experiments on Olivetti Face Database}

\begin{figure}[!bht]
\centering
\includegraphics[width=0.9\linewidth]{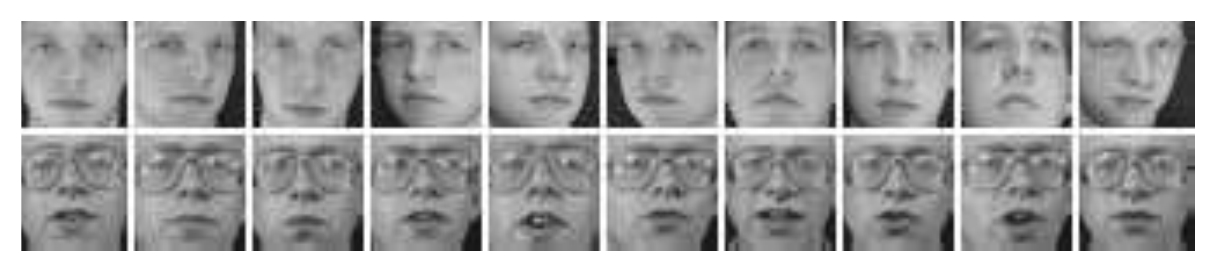}
\caption{The first two categories of Oliv.-40. Images of the first person vary with the angle to the camera,
and images of the second person vary greatly with the facial expressions.
}\label{fig:olivetti-faces-examples}
\end{figure}

The last real world data set is the Olivetti Face Database, served for the task of unsupervised face clustering.
The Olivetti Face Database is formerly the ORL Database of Faces, which consists of $400$ face images from $40$
individuals. The images are taken at different times, with varying lighting, facial expressions and facial
details. The size of each image is $64\times64$ pixels. Again, we select three subsets:
Oliv.-10 (faces $2,6,10,\cdots,34$ and $38$), Oliv.-20 (faces with the odd number),
and Oliv.-30 (Oliv.-10 + Oliv.-20). The whole Olivetti Face Databse is denoted as Oliv.-40.
Fig. \ref{fig:olivetti-faces-examples} presents some example faces of the Olivetti Face Database.

The manifold distance of the Oliv.-sets is approximated by using the graph distance with the CW-SSIM index and
the $3$-nn neighborhood.

%\text{~}

\subsubsection{Performance on the estimation of $k$}

It is a hard task to estimate $k$ for the Oliv.-sets due to the limited samples in each category ($m_0=10$) and
high dimensionality ($p=4096$) but a relatively large number of clusters (total $k^*=40$).
TABLE \ref{tab:estimate-k-olivettifaces} listed the estimated $k$ of the comparing methods on the Oliv.-sets.
$x$-means, $dip$-means, CFSFDP(E) and LDPS(E) all fail to learn a reasonable $k$. The estimated $k$ of Oliv.-40
by CFSFDP with the CW-SSIM index is around $30$ \cite{rodriguez2014clustering}, much close to $k^*$ compared with
the previous methods. CFSFDP(M) learns a reasonable $k$ for Oliv.-10 and Oliv.-20, but badly underestimated
$k$ for Oliv.-30 and Oliv.-40. LDPS(M) gets the consistent cluster number with the ground truth
for the first three Oliv.-sets. Though it underestimated the number of clusters on Oliv.-40,
its estimated result is much closer to the ground truth than that obtained by the other comparing methods.

\subsubsection{Clustering performance}

TABLE \ref{tab:clustering-olivettifaces} shows the clustering results on the Oliv.-sets.
$k$-means and LDPS-means obtained bad results on the last three Oliv.-sets due to their use of the Euclidean distance
as the dissimilarity measure.
%Relatively, LDPS-means gets a little bit better results than $k$-means.
With the help of the manifold distance, $k$-medoids and LDPS-medoids obtained much better results than $k$-means
and LDPS-means did. LDPS-medoids outperforms $k$-medoids with the properly selected initial seeds. In addition,
when setting $k^*$ to be $42$ for Oliv.-40, LDPS-medoids gets $r_e=18.5\%$, $r_t=74.0\%$ and $r_f=0.9\%$.
This improves $r_e$ by $15.9\%$ decreasing, $r_t$ by $8.8\%$ increasing and $r_f$ by $25\%$ decreasing over
the results reported in \cite{rodriguez2014clustering}, where $r_e=22.0\%$, $r_t=68\%$ and $r_f=1.2\%$.

%=======================================================
%                   conclusion
%=======================================================
\section{Conclusion and Future Works} \label{sec:conclusion}

In this paper, we proposed a novel method, the LDPS algorithm, to learn the appropriate number
of clusters and to select deterministically the initial seeds of clusters with high quality for the $k$-means-like methods.
In addition, two novel methods, LDPS-means and LDPS-medoids, have also been proposed as a combination of the LDPS
initialization algorithm and the $k$-means-like clustering algorithm.
Performance analysis and experimental results have demonstrated that our methods have the following advantages:
\begin{enumerate}
\item The LDPS algorithm can learn a reasonable number $k$ of clusters for data sets with balanced samples in each category.
In addition, it can deal with a variety of data distributions with an appropriate dissimilarity measure.
\item The initial seeds selected by LDPS-means/LDPS-medoids are geometrically close to the centers of the clusters.
As a result, our methods can achieve a very good $\text{SSE}^*$, which is sometimes competitive with that achieved
by $k$-means/$k$-medoids with thousands of repeats. In addition, the number of iterations in the clustering stage of
our methods is generally much less than that needed by the other methods that select the initial seeds randomly.
\item Our methods give superior results in dealing with data sets with very large true cluster number $k^*$,
compared with $k$-means/$k$-medoids. This is mainly due to the local distinctiveness index introduced in this paper.
\item LDPS-medoids gives a superior performance on the unsupervised object clustering tasks. This is mainly due to the use
of the LDPS algorithm for deterministic initialization and the manifold distance (based on the CW-SSIM index and the
$t$-nn neighborhood) as the dissimilarity measure.
\end{enumerate}

Despite the above advantages, our methods have also some limitations. First, the time complexity of our methods is relatively higher compared with the original $k$-means algorithm. Thus, our methods can not deal with very large data sets. Secondly, difficulty may occur in parameters estimation when dealing with unbalanced data sets. In addition, we did not give detailed discussion on the outliers issue. These issues will be considered in the future.

% \section*{Acknowledgment}

% This work

\bibliographystyle{IEEEtran}
\bibliography{IEEEabrv,ldps_ref}

\appendices
%=======================================================
%                   append1
%=======================================================
\section{Proof of Theorem 1} \label{append1}

To prove the theorem, we need two basic results in mathematical analysis \cite{rudin1964principles}.

\newtheorem{lemma}{Lemma}
\begin{lemma}
$\lim\limits_{n \rightarrow \infty} \frac{(n!)^{\frac{1}{n}}}{n} = \frac{1}{e}$.
\end{lemma}

\begin{lemma}
$\lim\limits_{n \rightarrow \infty} (1+\frac{1}{n})^{n} = e$.
\end{lemma}

\def\proof{\noindent{\it Proof of Theorem 1: }}
\begin{proof}
First, local density peaks would geometrically near the center of the clusters as analyzed in sec. \ref{subsec:ldps} under conditions 1)-3). With the local density peaks as initial seeds, LDPS-means could separate the clusters with only O(1) iterations. Second, the $k$-means achieves the competitive performance when the initial seeds are selected with each cluster a seed. The statistical event that the $k$-means could achieve this in one repeat forms a Bernoulli distribution \cite{papoulis2002probability}, with probability
$$P(\text{success}) \overset{\text{\tiny cond. 2)}}{=} m_0^k / \mychoose{m}{k}.$$
In addition, the distribution of $P(\text{success})$ between two different repeats are independent and identical. As a consequence, the expected number of repeats (\#repeats) to achieve the competitive performance is,
\begin{equation}
E(\text{\#repeats})= \frac{1}{P(\text{success})} = \mychoose{m}{k} / m_0^k\, .
\label{eq:repeats-estimate}
\end{equation}
Finally, by the lemmas and conditions, we get
\begin{eqnarray}
\mychoose{m}{k} & = & \frac{m!}{(m-k)! \cdot k!} \nonumber \\
 & \overset{\text{\tiny lem. 1}}{=} & O\big(\frac{(m/e)^m}{((m-k)/e)^{m-k} \cdot (k/e)^k }\big) \nonumber \\
 & = & O\big( (\frac{m}{m-k})^{m-k} \cdot (\frac{m}{k})^k\big) \nonumber \\
 & \overset{\text{\tiny cond. 2)}}{=} & O\big((1+\frac{1}{m_0-1})^{(m_0-1)k} \cdot m_0^k \big) \nonumber \\
 & \overset{\text{\tiny  lem. 2}}{=} & O\big(e^k \cdot m_0^k \big) \label{eq:choose}
\end{eqnarray}
Thus, with (\ref{eq:repeats-estimate}) and (\ref{eq:choose}) we get $E(\text{\#repeats}) = O(e^k)$. \qed
\end{proof}

\end{document}